\documentclass[final]{siamltex}
\usepackage{graphicx} 
\usepackage{mathtools}
 
\usepackage{latexsym}
 \usepackage{booktabs}
\usepackage{tabularx}
\usepackage{ragged2e}
\usepackage[table]{xcolor}
\usepackage[mathscr]{euscript}
\usepackage{algorithm,algorithmic}

\usepackage{amssymb}
\usepackage{fancyhdr}
\usepackage{multirow}
\usepackage{epsfig}
\usepackage{mathptmx}       
\usepackage[most]{tcolorbox}
\usepackage{helvet}         
\usepackage{courier}        
\usepackage{type1cm}        
\usepackage{imakeidx}         
\usepackage[a4paper,textheight=23cm,textwidth=15cm]{geometry} 
\usepackage{hyperref}
\usepackage{enumitem}
\usepackage{titlesec}
\usepackage{amsmath,amssymb,amsfonts}
\usepackage{tikz}
\usepackage{booktabs}
\usepackage{bm}
\usetikzlibrary{arrows.meta,shapes,positioning,calc,fit,backgrounds,decorations.pathreplacing}
\usepackage{pifont}   

\newtheorem{remark}{remark}

\title{Structured Multidimensional Representation Learning for Large Language Models }
\author{A. Elichi\thanks{Université du Littoral Cote d'Opale, LMPA, 50 rue F. Buisson, 62228 Calais-Cedex, France.} \and K. Jbilou\footnotemark[1]  \and M. El Guide \thanks{FGSES, University Mohammed VI Polytechnic, Rabat, Morocco}    \and  F. Dufrenois\thanks{ LISIC, 50 rue F. Buisson, Universit\'e du Littoral Cote d'Opale, 62228 Calais-Cedex, France.} }
\date{}

\begin{document}

\maketitle
\begin{abstract}
Transformer architectures achieve state-of-the-art performance across a wide range of pattern recognition and natural language processing tasks, but their scaling is accompanied by substantial parameter growth and redundancy in the embedding dimension. In this work, we introduce a structured spectral factorization of the embedding space based on the $\mathcal{L}$-product for third-order tensors. By reshaping token representations into spectral tensor slices and performing attention and feed-forward operations in the transform domain, we obtain a \emph{Tensor Transformer} architecture that decomposes the encoder into $p$ independent spectral sub-transformers while preserving standard Transformer semantics. We prove that the proposed $\mathcal{L}$-Transformer is spectrally equivalent to $p$ parallel Transformers operating on reduced-dimensional embeddings, which yields an $\approx 1/p$ reduction (up to lower-order terms such as biases and normalization parameters) in encoder parameters under fixed total embedding size. When instantiated with a real-valued Discrete Cosine Transform (DCT), the method remains fully differentiable and compatible with existing training pipelines. Beyond compression, the spectral decomposition introduces an inductive bias over embedding frequencies, enabling slice-dependent frequency scaling that improves generalization. Experiments on IMDB and AG~News show that the proposed model can substantially reduce
encoder parameters (up to $75\%$ for $p{=}4$) while maintaining competitive accuracy.
On IMDB, the tensorized encoder matches or improves upon the standard baseline under
compression, whereas on AG~News at moderate width we observe a small accuracy decrease
in exchange for a $4\times$ encoder reduction; at BERT-base width ($d{=}768$), performance
returns to parity.
\end{abstract}
\section{Introduction}

Transformer architectures \cite{Vaswani2017} have become the dominant paradigm in modern pattern recognition and natural language processing, achieving state-of-the-art performance in text classification, machine translation, and large-scale language modeling \cite{Brown2020,Touvron2023,devlin2019bert,liu2019roberta,raffel2020exploring}. However, their success is accompanied by substantial growth in parameter count, particularly in the embedding dimension and feed-forward layers. As model width increases, redundancy and overparameterization emerge, motivating research on structured compression and low-rank approximations \cite{Yu2020,Sharma2024,Luo2024}, including distillation \cite{sanh2019distilbert,jiao2020tinybert}, architectural parameter sharing \cite{lan2019albert}, and structured factorizations \cite{tahaei2022kroneckerbert}.

Most existing compression strategies operate directly in the weight space through pruning, low-rank matrix factorization, or post-hoc rank reduction. While effective, these approaches approximate a pre-trained model and do not fundamentally alter the representation geometry of the embedding space. An alternative perspective consists in reparameterizing the embedding dimension itself through structured spectral factorization, thereby imposing algebraic structure prior to training.

Tensor algebra provides a natural framework for structured multidimensional representations \cite{KoldaBader2009,Oseledets2011,KilmerMartin2011}. In particular, the $\mathcal{L}$-product for third-order tensors defines multiplication via an invertible linear transform applied along a designated mode \cite{Kernfeld2015}. In the transform domain, this operation induces a block-diagonal structure that decouples tensor slices while preserving global consistency under inverse transformation.

In this work, we leverage this algebraic property to introduce a Tensor Transformer architecture. Token embeddings are reshaped into third-order tensors, and attention as well as feed-forward operations are reformulated using the $\mathcal{L}$-product. Our main theoretical result shows that the resulting encoder is spectrally equivalent to $p$ independent Transformers operating on reduced-dimensional embeddings in the transform domain. This equivalence yields an $\approx 1/p$ reduction in encoder parameters while maintaining the expressive capacity of standard self-attention within each spectral slice.

The proposed construction is not equivalent to a simple partition of the embedding dimension. The forward and inverse transforms couple spectral components through a global linear operator, allowing structured mixing across slices after each attention and feed-forward update. When instantiated with the Discrete Cosine Transform \cite{Ahmed1974DCT}, the framework remains real-valued, fully differentiable, and compatible with standard optimization pipelines.

In addition to parameter reduction, the spectral formulation introduces a frequency-domain inductive bias. By assigning slice-dependent positional scaling coefficients, the model can emphasize low-frequency components or distribute attention harmonically across spectral channels. This structured spectral weighting can help task-dependent and stability in moderate-scale classification tasks.

Empirical evaluations on IMDB and AG News demonstrate that the Tensor Transformer substantially reduce
encoder parameters. In certain configurations, spectral weighting further improves predictive performance relative to standard Transformer baselines. These findings suggest that structured spectral factorization provides a principled and computationally efficient alternative to flat embedding representations in attention-based models.

Section~\ref{sec:related} reviews related work on efficient Transformers, low-rank adaptation, tensorized parameterizations, and transform-based mixing.
Section~\ref{sec:prelim} introduces tensor notation and the algebraic operations needed in the sequel.
Section~\ref{sec:Lproduct} defines the $\mathcal{L}$-transform and $\mathcal{L}$-product, and recalls key properties such as slice-wise multiplication and the $\mathcal{L}$-SVD.
Section~\ref{sec:tensor_llm} presents the proposed tensorized Transformer architecture: the embedding tensorization, $\mathcal{L}$-multi-head attention, $\mathcal{L}$-FFN, normalization, and the slice-wise equivalence result underpinning the $\approx 1/p$ encoder-parameter scaling.
Section~\ref{sec:experiments} reports empirical results and ablations on IMDB and AG~News, including parameter-matched baselines, scaling to $d=768$, and efficiency measurements.
Finally, Section~\ref{sec:conclusion} concludes and outlines limitations and future directions, while the appendix provides additional implementation details and supplementary analyses.
\section{Related Work}
\label{sec:related}
Transformers have become the dominant architecture for sequence modelling by combining multi-head self-attention
with position-wise feed-forward blocks, enabling scalable representation learning across language, vision, and
multimodal settings \cite{Vaswani2017}. Scaling this architecture has also produced foundation language models
that achieve strong few-shot performance and broad generalization \cite{Brown2020,Touvron2023,devlin2019bert,liu2019roberta,raffel2020exploring}. At the same time, standard self-attention can be memory intensive and exhibits structural redundancies (including redundancy across heads), motivating a large body of work on approximating token interactions and compressing parameters while preserving downstream accuracy \cite{Yu2020,kitaev2020reformer,wang2020linformer,choromanski2020rethinking,dao2022flashattention}. These efforts range from architectural modifications that alter how
tokens interact, to post-training interventions that directly modify weight matrices, as well as parameter-efficient
fine-tuning strategies that avoid updating the full model.

A prominent family of approaches targets redundancy through low-rank structure. In parameter-efficient fine-tuning,
low-rank adapters inject trainable rank-constrained updates into pretrained linear maps, yielding strong accuracy at
low update cost \cite{Hu2021LoRA}. Complementary post-training methods compress or modify pretrained weights without
additional training. Layer-selective rank reduction removes higher-order components of selected weight matrices and
can improve downstream reasoning without adding parameters or requiring extra data \cite{Sharma2024}. Related
tensor-centric compression ideas can also be applied directly to collections of matrices arising in large models.
TRAWL represents model parameters in a tensor format and performs structured reduction to lower storage and
computation costs \cite{Luo2025TRAWL_PAKDD}. Together, these works motivate treating groups of related matrices as
structured objects rather than independent 2D arrays, which is especially natural in multi-head attention where heads
form an additional mode with meaningful correlations.  Complementary compression lines include knowledge distillation for pretrained Transformers \cite{sanh2019distilbert,jiao2020tinybert}, architectural parameter sharing \cite{lan2019albert}, and structured matrix factorizations tailored to Transformer blocks \cite{tahaei2022kroneckerbert}.

A second line of research develops tensor decompositions and tensorized network parameterizations to reduce complexity
by exploiting multilinear structure. Survey work has established CP, Tucker, and related decompositions as central
tools for compressing and regularizing high-dimensional models \cite{KoldaBader2009,tucker1966some,hitchcock1927expression}.
The tensor-train representation is particularly attractive for very high-dimensional tensors due to its favourable scaling
and stable algebra \cite{Oseledets2011}, and it has been used to tensorize neural network layers so that large weight matrices
are stored and manipulated in a compact factorized form \cite{Novikov2015Tensorizing}. In the context of Transformers,
these perspectives support viewing the parameters and activations in attention as higher-order tensors, which can be manipulated
with multilinear operators to capture inter-mode dependencies that are obscured by flattening.

Within tensor algebra, an influential framework is the transform-based tensor--tensor product that treats a third-order tensor
as a linear operator acting on matrices. The t-product formalism and its associated tensor SVD provide an operator-theoretic
analogue of matrix factorization that is well suited for modelling correlations along a designated tensor mode \cite{KilmerMartin2011}.
This perspective was generalized by defining tensor--tensor products induced by an invertible linear transform applied along the tube
dimension, yielding a flexible family of products that includes Fourier-type constructions while allowing other orthogonal transforms
\cite{Kernfeld2015}. Such transform-based products are appealing in learning settings because they combine two properties that are useful
for model design: they preserve a matrix-like calculus slice-wise in the transform domain, and they provide a principled way to encode
cross-slice interactions through the chosen transform.

Finally, spectral mixing has also appeared as an efficient alternative to explicit attention, replacing learned token-to-token interaction
kernels with fixed transforms that mix information globally with low overhead. FNet, for example, replaces attention with Fourier mixing
layers and demonstrates competitive results in several regimes \cite{LeeThorp2021FNet}. This connects naturally to approaches that leverage
real orthogonal transforms such as the discrete cosine transform, which can offer similar global mixing while remaining purely real valued.
Taken together, low-rank adaptation and post-training rank reduction, tensorized parameterizations and weight compression, transform-based
tensor--tensor products, and spectral mixing layers provide a coherent foundation for treating attention and its parameters as structured tensors
and for designing transform-domain operators that capture cross-head dependencies efficiently. The following subsection summarizes these axes and
clarifies where the proposed L-product tensorization fits.

\paragraph{Positioning of the proposed L-product tensorization among efficient Transformer methods}
\label{sec:positioning}

Work on efficient Transformers can be grouped by (i) what object is structured (weights versus representations) and (ii) when the structure
is imposed (during training versus post hoc). Parameter-efficient fine-tuning methods such as LoRA inject low-rank updates into pretrained
linear maps, reducing the number of trainable parameters while leaving the base model unchanged \cite{Hu2021LoRA}. Complementary post-training
approaches, such as layer-selective rank reduction, modify pretrained weights without additional task training \cite{Sharma2024}. A third family
compresses weights by representing collections of matrices through tensor formats or decompositions; recent work such as TRAWL applies structured
tensor reduction to large-model weights \cite{Luo2025TRAWL_PAKDD}, while classical foundations include CP/Tucker and tensor-train representations
\cite{KoldaBader2009,tucker1966some,hitchcock1927expression,Oseledets2011} and earlier tensorization strategies for neural networks
\cite{Novikov2015Tensorizing}. Finally, spectral mixing layers replace learned token interaction with fixed transforms; FNet replaces attention with
Fourier mixing \cite{LeeThorp2021FNet}.

The present work differs primarily in where the structure lives: it reparameterizes the embedding and intermediate representations through an
L-product induced by an invertible transform \cite{Kernfeld2015}, rather than approximating individual weight matrices. In the transform domain,
the resulting encoder is spectrally equivalent to $p$ independent Transformers operating on reduced-dimensional embeddings, yielding an exact $1/p$
reduction of encoder parameters under fixed total embedding size (while shared embeddings and task heads remain unchanged). This equivalence is
accompanied by global coupling after each block through the inverse transform, so the construction is not a simple feature partition.

\begin{table}[h!]
\centering
\caption{Closest-method comparison and positioning of the proposed $\mathcal{L}$-product tensorization.}
\label{tab:closest_methods}
\setlength{\tabcolsep}{5pt}
\renewcommand{\arraystretch}{1.15}
\small
\begin{tabularx}{\textwidth}{@{}>{\RaggedRight\arraybackslash}p{2.9cm}
                        >{\RaggedRight\arraybackslash}p{2.8cm}
                        >{\RaggedRight\arraybackslash}p{2.0cm}
                        >{\RaggedRight\arraybackslash}p{2.6cm}
                        >{\RaggedRight\arraybackslash}X@{}}
\toprule
\textbf{Method family} &
\textbf{What is structured} &
\textbf{When applied} &
\textbf{Exact vs.\ approximate} &
\textbf{Primary gain / typical limitation} \\
\midrule
Low-rank PEFT (LoRA) \cite{Hu2021LoRA} &
Weight updates &
Fine-tune &
Approx.\ to full fine-tuning &
Few trainable parameters; base encoder parameter count unchanged. \\

Post-training rank reduction \cite{Sharma2024} &
Weights &
Post hoc &
Approx.\ modified model &
No retraining; accuracy depends on layer selection and rank choice. \\

Tensorized weight compression (TRAWL; CP/Tucker/TT)
\cite{Luo2025TRAWL_PAKDD,KoldaBader2009,Oseledets2011,tucker1966some,hitchcock1927expression} &
Weights as tensor factors &
Train or post hoc &
Approx.\ factorization &
Storage/parameters reduced; may affect optimization and accuracy. \\

Spectral mixing (FNet-like) \cite{LeeThorp2021FNet} &
Token mixing operator &
Train &
Exact transform (different layer type) &
Global mixing at low parameter cost; not targeted at exact $1/p$ encoder scaling. \\

\rowcolor{gray!12}
\textbf{This work ($\mathcal{L}$-product tensorization)} \cite{Kernfeld2015} &
\textbf{Representations and tensor operators} &
\textbf{Train-time reparameterization} &
\textbf{Exact slice-wise equivalence (transform domain)} &
\textbf{Encoder parameters scale as $\approx 1/p$; attention-map storage and the $T^2 d$ attention term remain unchanged; wall-clock gains require batched slice execution.} \\
\bottomrule
\end{tabularx}
\end{table}

\section{Preliminaries and notation}
\label{sec:prelim}

\subsection{Tensors, entries, slices, and fibers}
A tensor is a multidimensional array. The \emph{order} (also called the number of \emph{modes} or \emph{ways}) is the number of indices needed to address an entry. Scalars, vectors, and matrices correspond respectively to tensors of order $0$, $1$, and $2$.

Let $\mathcal{A}\in\mathbb{R}^{n_1\times n_2\times \cdots \times n_N}$ be an order-$N$ tensor. Its entries are denoted by
\[
a_{i_1 i_2 \cdots i_N}, \qquad 1\le i_k \le n_k,\; k=1,\dots,N.
\]
The Frobenius norm is
\[
\|\mathcal{A}\|_F^2
=\sum_{i_1=1}^{n_1}\sum_{i_2=1}^{n_2}\cdots\sum_{i_N=1}^{n_N} a_{i_1 i_2 \cdots i_N}^2.
\]

In this work, third-order tensors are central. For $\mathcal{A}\in\mathbb{R}^{n_1\times n_2\times n_3}$:
\begin{itemize}
\item A \emph{frontal slice} is the matrix $\mathcal{A}(:,:,k)\in\mathbb{R}^{n_1\times n_2}$.
\item A \emph{fiber} is obtained by fixing all indices but one.
\item A \emph{tube} is a mode-$3$ fiber, i.e., $\mathcal{A}(i,j,:)\in\mathbb{R}^{1\times 1\times n_3}$.
\end{itemize}

\subsection{Matricization (unfolding)}
The mode-$n$ matricization (unfolding) of $\mathcal{A}\in\mathbb{R}^{n_1\times\cdots\times n_N}$ is denoted by $\mathcal{A}_{(n)}$ and belongs to
\[
\mathcal{A}_{(n)} \in \mathbb{R}^{n_n \times \prod_{k\ne n} n_k }.
\]
An explicit index mapping  is placed in  Appendix~\ref{app:unfolding-conv}.

\subsection{$n$-mode product}
The $n$-mode product defines the multiplication of a tensor by a matrix along mode $n$.

\begin{definition}[$n$-mode product]
Let $\mathcal{A}=[a_{i_1 i_2 \cdots i_N}]\in\mathbb{R}^{I_1\times\cdots\times I_N}$ and
$X=[x_{j i_n}]\in\mathbb{R}^{J\times I_n}$.
The $n$-mode product $\mathcal{A}\times_n X$ is the order-$N$ tensor of size
\[
I_1\times\cdots\times I_{n-1}\times J \times I_{n+1}\times\cdots\times I_N,
\]
with entries
\[
(\mathcal{A}\times_n X)_{i_1\cdots i_{n-1} j i_{n+1}\cdots i_N}
= \sum_{i_n=1}^{I_n} a_{i_1\cdots i_N}\, x_{j i_n}.
\]
\end{definition}

Equivalently, in terms of unfolding,
\[
\mathcal{B}=\mathcal{A}\times_n X
\quad\Longleftrightarrow\quad
\mathcal{B}_{(n)} = X\,\mathcal{A}_{(n)}.
\]

Similarly, the contraction with a vector $v\in\mathbb{R}^{I_n}$ along mode $n$ yields an order-$(N-1)$ tensor, denoted here by $\mathcal{A}\,\bar{\times}_n v$ \footnote{We use the overbar to explicitly denote the order-reduction step}, defined by
\[
(\mathcal{A}\,\bar{\times}_n v)_{i_1\cdots i_{n-1} i_{n+1}\cdots i_N}
= \sum_{i_n=1}^{I_n} a_{i_1\cdots i_N}\, v_{i_n}.
\]

\begin{remark}
Many standard tensor decompositions can be written compactly using the $n$-mode product, including the CANDECOMP/PARAFAC (CP) decomposition \cite{hitchcock1927expression}, the Tucker decomposition \cite{tucker1966some}, and the Tensor Train (TT) decomposition \cite{Oseledets2011}.
\end{remark}

\section{The $\mathcal{L}$-Product}
\label{sec:Lproduct}

We review the $\mathcal{L}$-product (a transform-based tensor multiplication) which unifies several tensor algebra frameworks through slice-wise operations in a transform domain \cite{Kernfeld2015}.

\subsection{$\mathcal{L}$-transform and facewise multiplication}

\begin{definition}[$\mathcal{L}$-transform]\label{def:L-transform}
Let $Z\in\mathbb{R}^{p\times p}$ be invertible. Define the linear transform
\[
\mathcal{L}:\mathbb{R}^{n\times m\times p}\to\mathbb{R}^{n\times m\times p},
\qquad
\mathcal{L}(\mathcal{A}) = \mathcal{A}\times_3 Z,
\]
with inverse
\[
\mathcal{L}^{-1}(\mathcal{A}) = \mathcal{A}\times_3 Z^{-1}.
\]
\end{definition}

We use the shorthand
\[
\widehat{\mathcal{A}} := \mathcal{L}(\mathcal{A}),
\qquad
\widehat{A}^{(k)} := \widehat{\mathcal{A}}(:,:,k),\; k=1,\dots,p.
\]

\begin{definition}[Facewise product]
Let $\widehat{\mathcal{A}}\in\mathbb{R}^{m\times \ell\times p}$ and
$\widehat{\mathcal{B}}\in\mathbb{R}^{\ell\times n\times p}$.
Their \emph{facewise product} $\widehat{\mathcal{C}}=\widehat{\mathcal{A}}\triangle \widehat{\mathcal{B}}$
is the tensor in $\mathbb{R}^{m\times n\times p}$ whose frontal slices satisfy
\[
\widehat{C}^{(k)} = \widehat{A}^{(k)}\,\widehat{B}^{(k)},
\qquad k=1,\dots,p.
\]
\end{definition}

\begin{definition}[$\mathcal{L}$-product]
Let $\mathcal{A}\in\mathbb{R}^{m\times \ell\times p}$ and
$\mathcal{B}\in\mathbb{R}^{\ell\times n\times p}$.
Their $\mathcal{L}$-product is
\begin{equation}\label{eq:L-product}
\mathcal{A}*_{\mathcal{L}}\mathcal{B}
:= \mathcal{L}^{-1}\!\Big(\widehat{\mathcal{A}}\triangle \widehat{\mathcal{B}}\Big)
= \mathcal{L}^{-1}\!\Big(\mathcal{L}(\mathcal{A})\triangle \mathcal{L}(\mathcal{B})\Big).
\end{equation}
\end{definition}

\subsection{Identity, transpose, orthogonality, and invertibility}

\begin{definition}[$\mathcal{L}$-identity tensor]
The identity tensor $\mathcal{I}\in\mathbb{R}^{m\times m\times p}$ is defined by requiring
\[
\widehat{\mathcal{I}}(:,:,k)=I_m,\qquad k=1,\dots,p,
\]
and setting $\mathcal{I}:=\mathcal{L}^{-1}(\widehat{\mathcal{I}})$.
\end{definition}

\begin{definition}[$\mathcal{L}$-transpose]
Let $\mathcal{A}\in\mathbb{R}^{m\times n\times p}$. Its $\mathcal{L}$-transpose
$\mathcal{A}^T\in\mathbb{R}^{n\times m\times p}$ is defined by
\[
\widehat{(\mathcal{A}^T)}(:,:,k) = \big(\widehat{\mathcal{A}}(:,:,k)\big)^{H},
\qquad k=1,\dots,p,
\]
where $(\cdot)^H$ denotes conjugate transpose (which reduces to transpose in the real-valued case).
\end{definition}

\begin{definition}[Structured tensors under the $\mathcal{L}$-product]
A square tensor $\mathcal{Q}\in\mathbb{R}^{m\times m\times p}$ is:
\begin{itemize}
\item \emph{$\mathcal{L}$-orthogonal} if
\[
\mathcal{Q}^T*_{\mathcal{L}}\mathcal{Q}
=\mathcal{Q}*_{\mathcal{L}}\mathcal{Q}^T
=\mathcal{I}.
\]
Equivalently, every frontal slice $\widehat{Q}^{(k)}$ is unitary/orthogonal.
\item \emph{f-diagonal} if each frontal slice $\widehat{Q}^{(k)}$ is diagonal.
\item \emph{invertible} if there exists $\mathcal{Q}^{-1}\in\mathbb{R}^{m\times m\times p}$ such that
\[
\mathcal{Q}*_{\mathcal{L}}\mathcal{Q}^{-1}
=\mathcal{Q}^{-1}*_{\mathcal{L}}\mathcal{Q}
=\mathcal{I}.
\]
Equivalently, every $\widehat{Q}^{(k)}$ is invertible.
\end{itemize}
\end{definition}

\subsection{The $\mathcal{L}$-SVD and ranks}
\medskip

\begin{theorem}[$\mathcal{L}$-SVD]\label{thm:L-SVD}
Let $\mathcal{A}\in\mathbb{R}^{m\times n\times p}$. Then there exist $\mathcal{L}$-orthogonal tensors
$\mathcal{U}\in\mathbb{R}^{m\times m\times p}$ and $\mathcal{V}\in\mathbb{R}^{n\times n\times p}$,
and an f-diagonal tensor $\mathcal{S}\in\mathbb{R}^{m\times n\times p}$ such that
\[
\mathcal{A}=\mathcal{U}*_{\mathcal{L}}\mathcal{S}*_{\mathcal{L}}\mathcal{V}^T.
\]
The diagonal tubes $\mathcal{S}(i,i,:)$ are called \emph{singular tubes}; their $\ell_2$-norms play the role of singular values in the $\mathcal{L}$-framework \cite{KilmerMartin2011, kilmer2013third}.
\end{theorem}

\begin{remark}[Computation]
In practice, one computes $\widehat{\mathcal{A}}=\mathcal{L}(\mathcal{A})$, performs $p$ independent matrix SVDs
\(
\widehat{A}^{(k)} = \widehat{U}^{(k)} \widehat{S}^{(k)} (\widehat{V}^{(k)})^{H},
\)
and maps back by $\mathcal{L}^{-1}$ to obtain $\mathcal{U},\mathcal{S},\mathcal{V}$.
\end{remark}

\begin{definition}[$\mathcal{L}$-average rank]
For $\mathcal{A}\in\mathbb{R}^{m\times n\times p}$, the $\mathcal{L}$-average rank is
\[
\operatorname{rank}_a(\mathcal{A})
:=\frac{1}{p}\sum_{k=1}^{p}\operatorname{rank}\!\big(\widehat{A}^{(k)}\big).
\]
\end{definition}

\begin{definition}[$\mathcal{L}$-tubal rank]\label{def:tubal-rank}
Let $\mathcal{A}=\mathcal{U}*_{\mathcal{L}}\mathcal{S}*_{\mathcal{L}}\mathcal{V}^T$ be an $\mathcal{L}$-SVD.
The $\mathcal{L}$-tubal rank is the number of nonzero singular tubes:
\[
\operatorname{rank}_T(\mathcal{A})
:= \#\Big\{ i \in \{1,\dots,\min(m,n)\} \;:\; \|\mathcal{S}(i,i,:)\|_2 > 0 \Big\}.
\]
(For noisy data, one typically uses a small threshold instead of $0$.)
\end{definition}

\subsection{Truncated $\mathcal{L}$-SVD (spectral truncation)}
Often, only a few dominant singular tubes are needed. A truncated $\mathcal{L}$-SVD keeps the $k$ largest singular tubes (by $\ell_2$-norm) and yields
\[
\mathcal{A}\approx \mathcal{U}_k*_{\mathcal{L}}\mathcal{S}_k*_{\mathcal{L}}\mathcal{V}_k^T,
\qquad k\ll \min(m,n).
\]
This spectral truncation reduces effective degrees of freedom, improves numerical stability, and can substantially reduce storage and compute costs.

\subsection{Special cases: t-product and DCT-based products}
Several classical tensor products appear as special cases of the $\mathcal{L}$-product \cite{Kernfeld2015}:
\begin{itemize}
\item \textbf{t-product:} choose $Z$ as the (unitary) discrete Fourier transform matrix; then $\mathcal{L}$ corresponds to applying the FFT along tubes, yielding complex-valued transform slices.
\item \textbf{DCT-based products:} choose $Z$ as a discrete cosine transform matrix \cite{Ahmed1974DCT}; then $\mathcal{L}$ uses DCT along tubes and remains real-valued for real data.
\end{itemize}

\begin{remark}
FFT-based transforms generally produce complex coefficients (magnitude and phase), while DCT-based transforms are purely real for real inputs.
Moreover, DCT often exhibits strong energy compaction \cite{Ahmed1974DCT}, which can be advantageous for compression, denoising by spectral truncation, and low-rank tensor approximation.
\end{remark}
\section{Tensor LLMs under the $\mathcal{L}$-product framework}
\label{sec:tensor_llm}

A \emph{Tensor Large Language Model (Tensor LLM)} is a Transformer-style neural network in which internal representations and parameters are organized as higher-order tensors and combined through the $\mathcal{L}$-product algebra.
Compared to standard Transformers that rely on matrix multiplication, Tensor LLMs operate in a transform-induced tensor space where multilinear structure can be explicitly exploited.

\subsection{Tensorization of token embeddings}
\label{subsec:tensorization}

Let $X \in \mathbb{R}^{T \times d}$ be the usual input sequence (length $T$, embedding dimension $d$).
Fix an integer $p$ such that $p \mid d$ and define the \emph{slice width} $d_s := d/p$. \footnote{The clean tensorization assumes $p$ divides $d$. If not, one may (i) choose a nearby admissible $p$, or (ii) pad the embedding dimension to the next multiple of $p$ and truncate back after unfolding; we keep $p\mid d$ throughout to avoid cluttering notation.}
We reshape $X$ into a third-order tensor
\[
\mathcal{X} = \operatorname{Ten}_p(X) \in \mathbb{R}^{T \times d_s \times p},
\]
by splitting the feature dimension into $p$ slices:
\begin{equation}
\label{eq:tensorize}
\mathcal{X}(t,j,k) := X\big(t,\,(k-1)d_s + j\big),
\quad
t=1,\dots,T,\; j=1,\dots,d_s,\; k=1,\dots,p.
\end{equation}
The inverse operation $\operatorname{Mat}_p(\cdot)$ concatenates the $p$ slices along mode-2 and recovers a matrix in $\mathbb{R}^{T\times d}$.

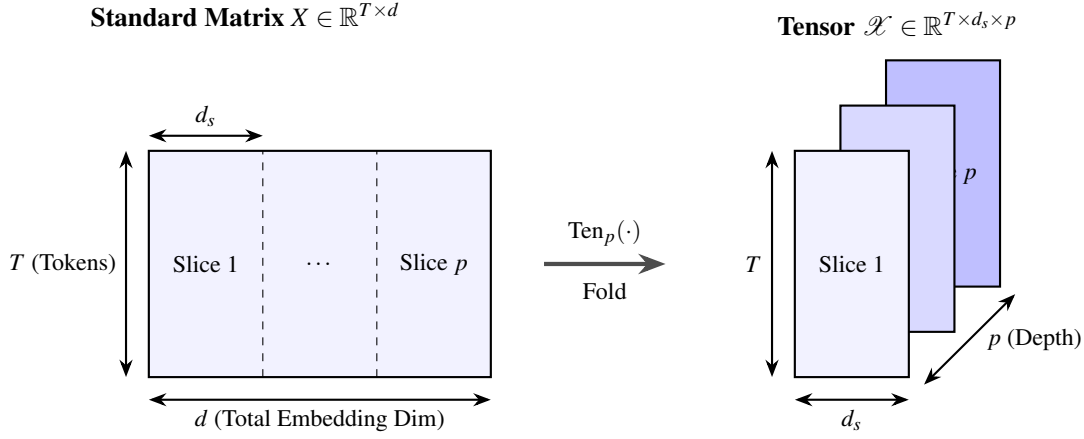
\begin{figure}[h!]
\centering
\begin{tikzpicture}[>=Stealth, font=\small]
    
    \begin{scope}[local bounding box=matrixBox]
        \draw[thick, fill=blue!5] (0,0) rectangle (4.5, 3);
        
        \draw[dashed, thin] (1.5,0) -- (1.5,3);
        \draw[dashed, thin] (3.0,0) -- (3.0,3);
        
        \draw[<->, thick] (-0.3, 0) -- (-0.3, 3) node[midway, left] {$T$ (Tokens)};
        \draw[<->, thick] (0, -0.3) -- (4.5, -0.3) node[midway, below] {$d$ (Total Embedding Dim)};
        
        \draw[<->, thick] (0, 3.2) -- (1.5, 3.2) node[midway, above] {$d_s$};
        \node at (0.75, 1.5) {Slice $1$};
        \node at (2.25, 1.5) {$\dots$};
        \node at (3.75, 1.5) {Slice $p$};
        
        \node[above=0.8cm of matrixBox.north, font=\bfseries] {Standard Matrix $X \in \mathbb{R}^{T \times d}$};
    \end{scope}

    \draw[->, ultra thick, draw=black!70] (5.2, 1.5) -- (6.8, 1.5) 
        node[midway, above=0.1cm] {$\operatorname{Ten}_p(\cdot)$}
        node[midway, below=0.1cm] {Fold};

    \begin{scope}[shift={(8.5, 0)}]
        
        \draw[thick, fill=blue!25] (1.2, 1.2) rectangle (2.7, 4.2);
        \node at (1.95, 2.7) {Slice $p$};
        
        \draw[thick, fill=blue!15] (0.6, 0.6) rectangle (2.1, 3.6);
        \node at (1.35, 2.1) {$\dots$};
        
        \draw[thick, fill=blue!5] (0,0) rectangle (1.5, 3);
        \node at (0.75, 1.5) {Slice $1$};
        
        \draw[<->, thick] (-0.3, 0) -- (-0.3, 3) node[midway, left] {$T$};
        \draw[<->, thick] (0, -0.3) -- (1.5, -0.3) node[midway, below] {$d_s$};
        \draw[<->, thick] (1.7, -0.1) -- (2.9, 1.1) node[midway, right=0.1cm] {$p$ (Depth)};
        
        \node[above=0.8cm of {(1.35,3.6)}, font=\bfseries] {Tensor $\mathcal{X} \in \mathbb{R}^{T \times d_s \times p}$};
    \end{scope}

\end{tikzpicture}
\caption{Tensorization of token embeddings. The matrix $X\in\mathbb{R}^{T\times d}$ is reshaped into
$\mathcal{X}\in\mathbb{R}^{T\times d_s\times p}$ with $d_s=d/p$ by splitting the feature dimension into
\emph{$p$ blocks of width $d_s$}. The third mode ($p$) is the tube dimension used by the $\mathcal{L}$-product;
applying $\mathcal{L}$ along mode-3 produces $p$ transform-domain frontal slices $\widehat{X}^{(k)}\in\mathbb{R}^{T\times d_s}$.}
\label{fig:tensor_folding}
\end{figure}

In what follows, the contextual representation will be produced in tensor form,
\[
\mathcal{Z} \in \mathbb{R}^{T \times d_s \times p},
\]
and can be unfolded back to the standard matrix representation $Z=\operatorname{Mat}_p(\mathcal{Z})\in\mathbb{R}^{T\times d}$ when needed.
\paragraph{From matrices to 3-way tensors.}
Given $X\in\mathbb{R}^{T\times d}$ and an integer $p$ such that $p\mid d$, we set $d_s=d/p$ and reshape
$X$ into a third-order tensor $\mathcal{X}\in\mathbb{R}^{T\times d_s\times p}$ by grouping the feature
dimension into \emph{$p$ contiguous blocks of width $d_s$}. Concretely, for each token position $t$,
the vector $X(t,:)\in\mathbb{R}^{d}$ is viewed as a $d_s\times p$ matrix, and the third mode index
$k\in\{1,\dots,p\}$ selects the \emph{slice/channel} while the second mode index $j\in\{1,\dots,d_s\}$
selects coordinates \emph{within} that slice.
In tensor terminology, each mode-3 fiber $\mathcal{X}(t,j,:)$ is a \emph{tube} of length $p$, and the
$\mathcal{L}$-transform is applied along this tube dimension.
Figure~\ref{fig:tensor_folding} provides a schematic view of this folding operation.

\subsection{$\mathcal{L}$-transform and spectral slicing}
\label{subsec:Ltransform_llm}

Let $Z \in \mathbb{R}^{p\times p}$ be an invertible transform matrix defining
\[
\widehat{\mathcal{A}}=\mathcal{L}(\mathcal{A})=\mathcal{A}\times_3 Z,
\qquad
\mathcal{L}^{-1}(\mathcal{A})=\mathcal{A}\times_3 Z^{-1}.
\]
We denote frontal slices in the transform domain by
\[
\widehat{A}^{(i)} := \widehat{\mathcal{A}}(:,:,i), \quad i=1,\dots,p.
\]
Throughout the implementation-oriented parts (LayerNorm/FFN), we focus on \emph{real orthogonal transforms} (notably DCT-type transforms), so that transform-domain slices remain real-valued for real inputs. \footnote{In all experiments we use an orthonormal DCT (DCT-II with orthonormal scaling), so $\mathcal{L}^{-1}$ is implemented by the corresponding inverse DCT with the same normalization. Other orthogonal transforms can be used without changing the slice-wise calculus; Fourier/DFT choices remain valid but lead to complex-valued slices (Appendix~\ref{app:complex}).}

\subsection{Basic operators: softmax and concatenation across heads}
\label{subsec:ops}

\begin{definition}[Row-wise softmax]
Let $S\in\mathbb{R}^{T\times T}$. The row-wise softmax is
\[
\mathrm{Softmax}(S)_{t,u}
=
\frac{\exp(S_{t,u})}{\sum_{v=1}^{T}\exp(S_{t,v})},
\quad t,u=1,\dots,T.
\]
Each row of $\mathrm{Softmax}(S)$ sums to $1$.
\end{definition}

\begin{definition}[Head concatenation within a slice]
\label{def:concat_slice}
Fix a transform-domain slice index $i\in\{1,\dots,p\}$ and let
$\widehat{H}^{(1)}_i,\dots,\widehat{H}^{(h)}_i \in \mathbb{R}^{T\times d_h}$
be the outputs of $h$ attention heads (with $d_s = h\,d_h$).
Define
\[
\mathrm{Concat}\big(\widehat{H}^{(1)}_i,\dots,\widehat{H}^{(h)}_i\big)
\in \mathbb{R}^{T\times d_s}
\]
as the column-wise concatenation.
\end{definition}

\subsection{Tensor positional encoding initialization}
\label{subsec:tensor_pe}

Let $\mathcal{X}\in\mathbb{R}^{T\times d_s\times p}$ be the tensorized input.
We define a slice-aware sinusoidal encoding $\mathcal{P}\in\mathbb{R}^{T\times d_s\times p}$ and set
\[
\mathcal{X}_{\text{pos}} = \mathcal{X} + \mathcal{P}.
\]

\begin{definition}[Slice-aware sinusoidal positional encoding]
\label{def:slice_sinusoidal_pe}
For $t=1,\dots,T$, $j=1,\dots,d_s$, $k=1,\dots,p$ define
\[
\mathcal{P}(t,j,k)=
\begin{cases}
\sin\!\Big(\dfrac{t}{10000^{\frac{2\lfloor (j-1)/2\rfloor}{d_s}}}\;\alpha_k\Big),
& \text{if $j$ is odd},\\[2mm]
\cos\!\Big(\dfrac{t}{10000^{\frac{2\lfloor (j-1)/2\rfloor}{d_s}}}\;\alpha_k\Big),
& \text{if $j$ is even},
\end{cases}
\]
where $\alpha_k$ is a slice-dependent frequency scaling factor.
\end{definition}

\begin{remark}[Frequency scaling strategies]
\label{def:alpha_strategies}
Common choices include:
\begin{enumerate}
\item \textbf{Linear (used here):} $\alpha_k = k/p$.
\item \textbf{Exponential:} $\alpha_k = 2^{(k-1)/(p-1)}$ (use $\alpha_1=1$ if $p=1$).
\item \textbf{Harmonic:} $\alpha_k = k$.
\end{enumerate}
\end{remark}
\paragraph{Learned absolute tensor PE (``learnable'' variant).}
In addition to fixed $\{\alpha_k\}$ strategies, we consider a \emph{learned} positional encoding where the entire tensor
\[
\mathcal{P}_{\text{learn}} \in \mathbb{R}^{T\times d_s\times p}
\]
is treated as trainable parameters and added once at the input:
$\mathcal{X}_{\text{pos}}=\mathcal{X}+\mathcal{P}_{\text{learn}}$.
This encoding is \emph{shared across all layers} (input-level absolute PE, as in standard learned position embeddings).
In our experiments it is regularized only through the same optimizer settings as the rest of the model (AdamW weight decay \cite{loshchilov2017decoupled}),
with no additional constraints (e.g., monotonicity) imposed.

\begin{remark}
When $p=1$ and $\alpha_1=1$, Definition~\ref{def:slice_sinusoidal_pe} reduces to the standard sinusoidal positional encoding.
\end{remark}

\subsection{$\mathcal{L}$-Multi-Head Attention}
\label{subsec:L_MHA}

We present tensor multi-head attention in a form that makes the slice-wise spectral computation explicit. 

\paragraph{Parameters.}
Let $h$ be the number of attention heads inside each slice, and set $d_s = h\,d_h$. \footnote{When comparing to a standard Transformer with $H$ heads at total width $d$, we allocate heads so that $H=p\,h$ and the per-head width matches: $d_h=d/H=d_s/h$. This keeps the attention granularity comparable across standard and tensorized encoders.}
For each head $j=1,\dots,h$, we use tensor-valued projections
\[
\mathcal{W}_Q^{(j)},\mathcal{W}_K^{(j)},\mathcal{W}_V^{(j)} \in \mathbb{R}^{d_s\times d_h\times p},
\qquad
\mathcal{W}_O \in \mathbb{R}^{d_s\times d_s\times p}.
\]
Their transform-domain versions are $\widehat{\mathcal{W}}_\bullet=\mathcal{L}(\mathcal{W}_\bullet)$.

\paragraph{Spectral-domain computation.}
Let $\widehat{\mathcal{X}}=\mathcal{L}(\mathcal{X}_{\mathrm{pos}})$ and denote $\widehat{X}^{(i)}=\widehat{\mathcal{X}}(:,:,i)\in\mathbb{R}^{T\times d_s}$.
For each slice $i=1,\dots,p$ and head $j=1,\dots,h$, define
\[
\widehat{Q}^{(j)}_i = \widehat{X}^{(i)}\,\widehat{W}^{(j)}_{Q,i},
\quad
\widehat{K}^{(j)}_i = \widehat{X}^{(i)}\,\widehat{W}^{(j)}_{K,i},
\quad
\widehat{V}^{(j)}_i = \widehat{X}^{(i)}\,\widehat{W}^{(j)}_{V,i},
\]
where $\widehat{W}^{(j)}_{Q,i} := \widehat{\mathcal{W}}_Q^{(j)}(:,:,i)$ (and similarly for $K,V$).
Scaled dot-product attention inside slice $i$ is
\[
\widehat{A}^{(j)}_i
=
\mathrm{Softmax}\!\left(\frac{\widehat{Q}^{(j)}_i(\widehat{K}^{(j)}_i)^\top}{\sqrt{d_h}}\right)
\in\mathbb{R}^{T\times T},
\qquad
\widehat{\mathrm{head}}^{(j)}_i
=
\widehat{A}^{(j)}_i\,\widehat{V}^{(j)}_i
\in\mathbb{R}^{T\times d_h}.
\]
Concatenate heads (Definition~\ref{def:concat_slice}) to obtain $\widehat{H}_i\in\mathbb{R}^{T\times d_s}$ and apply an output projection within slice $i$:
\[
\widehat{Y}_i = \widehat{H}_i\,\widehat{W}_{O,i},
\qquad \widehat{W}_{O,i} := \widehat{\mathcal{W}}_O(:,:,i).
\]
Stacking $\widehat{Y}_1,\dots,\widehat{Y}_p$ as frontal slices yields $\widehat{\mathcal{Y}}\in\mathbb{R}^{T\times d_s\times p}$, and the tensor output is
\[
\mathrm{MHA}_{\mathcal{L}}(\mathcal{X}_{\mathrm{pos}})
=
\mathcal{L}^{-1}(\widehat{\mathcal{Y}})\in\mathbb{R}^{T\times d_s\times p}.
\]

\begin{algorithm}[h]
\caption{$\mathcal{L}$-Multi-Head Attention (spectral implementation)}
\label{alg:mha}
\begin{algorithmic}[1]
\REQUIRE $\mathcal{X}\in\mathbb{R}^{T\times d_s\times p}$, $\mathcal{P}$, $\{\mathcal{W}_Q^{(j)},\mathcal{W}_K^{(j)},\mathcal{W}_V^{(j)}\}_{j=1}^h$, $\mathcal{W}_O$, transform $Z$
\ENSURE $\mathcal{Y}\in\mathbb{R}^{T\times d_s\times p}$
\STATE $\mathcal{X}_{\text{pos}}\gets \mathcal{X}+\mathcal{P}$
\STATE $\widehat{\mathcal{X}}\gets \mathcal{X}_{\text{pos}}\times_3 Z$
\STATE Transform all weights: $\widehat{\mathcal{W}}_\bullet \gets \mathcal{W}_\bullet\times_3 Z$
\FOR{$i=1$ to $p$ \textbf{in parallel}}
    \STATE $\widehat{X}^{(i)}\gets \widehat{\mathcal{X}}(:,:,i)$
    \FOR{$j=1$ to $h$}
        \STATE $\widehat{Q}_i^{(j)}\gets \widehat{X}^{(i)}\,\widehat{W}^{(j)}_{Q,i}$
        \STATE $\widehat{K}_i^{(j)}\gets \widehat{X}^{(i)}\,\widehat{W}^{(j)}_{K,i}$
        \STATE $\widehat{V}_i^{(j)}\gets \widehat{X}^{(i)}\,\widehat{W}^{(j)}_{V,i}$
        \STATE $\widehat{S}_i^{(j)}\gets \frac{1}{\sqrt{d_h}}\widehat{Q}_i^{(j)}(\widehat{K}_i^{(j)})^\top$
        \STATE $\widehat{\mathrm{head}}_i^{(j)}\gets \mathrm{Softmax}(\widehat{S}_i^{(j)})\,\widehat{V}_i^{(j)}$
    \ENDFOR
    \STATE $\widehat{H}_i \gets \mathrm{Concat}(\widehat{\mathrm{head}}_i^{(1)},\dots,\widehat{\mathrm{head}}_i^{(h)})$
    \STATE $\widehat{Y}_i \gets \widehat{H}_i\,\widehat{W}_{O,i}$
\ENDFOR
\STATE $\widehat{\mathcal{Y}} \gets \text{stack}(\widehat{Y}_1,\dots,\widehat{Y}_p)$
\STATE $\mathcal{Y}\gets \widehat{\mathcal{Y}}\times_3 Z^{-1}$
\RETURN $\mathcal{Y}$
\end{algorithmic}
\end{algorithm}
\paragraph{Implementation details}
Since $Z$ is fixed, we store the transformed weights $\widehat{\mathcal{W}}_{\bullet}
= \mathcal{W}_{\bullet}\times_3 Z$ as the trainable parameters directly and avoid
re-transforming weights at each forward pass; only activations are transformed. \footnote{In our implementation, activations are transformed once per encoder block (before slice-wise attention/FFN) and mapped back after the block. One can also keep activations in the transform domain across consecutive slice-wise sublayers to avoid repeated $\mathcal{L}/\mathcal{L}^{-1}$ calls; this is an engineering choice and does not change the slice-wise equivalence statements for the attention/FFN cores.}
\begin{remark}[Practical parallel implementation of the slice loop]
Although Algorithms~\ref{alg:mha}--\ref{alg:mha_detailed} are written with a loop ``for $i$ in parallel'',
in practice we implement slice-wise operations using \emph{batched} kernels.
A standard approach in PyTorch/JAX is to treat the slice index as a batch dimension by reshaping
$\widehat{\mathcal{X}}$ to shape $(p,T,d_s)$ (and fusing the head index so the effective batch is $p\cdot h$),
then using batched matrix multiplications / attention kernels in one call.
With this batching, all $p$ slices execute concurrently on GPU; if implemented as an explicit Python loop,
wall-clock time can scale close to linearly with $p$, which strongly affects timing results.
\end{remark}
A fully expanded slice/head-level implementation is given in Appendix~\ref{app:mha_detail}.

\subsection{$\mathcal{L}$-Feed-Forward Network}
\label{subsec:L_FFN}

\begin{definition}[Standard FFN (matrix)]
Let $X\in\mathbb{R}^{T\times d}$. A standard FFN is
\[
\mathrm{FFN}(X)=\sigma(XW_1+b_1)W_2+b_2,
\]
where $W_1\in\mathbb{R}^{d\times d_{ff}}$, $W_2\in\mathbb{R}^{d_{ff}\times d}$, and typically $d_{ff}=4d$.
\end{definition}

\begin{definition}[$\mathcal{L}$-FFN (tensor)]
Let $\mathcal{X}\in\mathbb{R}^{T\times d_s\times p}$ and set $d_{ff,s}=4d_s$.
Define tensor weights and biases
\[
\mathcal{W}_1\in\mathbb{R}^{d_s\times d_{ff,s}\times p},\quad
\mathcal{W}_2\in\mathbb{R}^{d_{ff,s}\times d_s\times p},\quad
\mathcal{B}_1\in\mathbb{R}^{1\times d_{ff,s}\times p},\quad
\mathcal{B}_2\in\mathbb{R}^{1\times d_s\times p}.
\]
Then
\begin{equation}
\label{eq:tffn}
\mathrm{TFFN}(\mathcal{X})
=
\sigma\!\big(\mathcal{X}*_{\mathcal{L}}\mathcal{W}_1+\mathcal{B}_1\big)
*_{\mathcal{L}}\mathcal{W}_2+\mathcal{B}_2,
\end{equation}
where $\sigma$ acts element-wise (e.g., ReLU or GELU).
\end{definition}

\begin{theorem}[$\mathcal{L}$-FFN in the transform domain]
\label{thm:ffn}
Let $\widehat{\mathcal{X}}=\mathcal{L}(\mathcal{X})$ and similarly for parameters.
Then for each slice $i=1,\dots,p$,
\[
\widehat{H}_i = \widehat{X}_i\,\widehat{W}_{1,i} + \widehat{b}_{1,i},
\quad
\widehat{G}_i = \sigma(\widehat{H}_i),
\quad
\widehat{Y}_i = \widehat{G}_i\,\widehat{W}_{2,i} + \widehat{b}_{2,i},
\]
which is exactly a standard FFN applied independently to each slice.
\end{theorem}

\subsection{Tensor layer normalization}
\label{subsec:layernorm}

\begin{definition}[Tensor LayerNorm]
Let $\mathcal{X}\in\mathbb{R}^{T\times d_s\times p}$.
Define mean and variance along the feature mode (mode-2) for each position $t$ and slice $k$:
\[
\mu_{t,k}=\frac{1}{d_s}\sum_{j=1}^{d_s}\mathcal{X}(t,j,k),
\qquad
\sigma^2_{t,k}=\frac{1}{d_s}\sum_{j=1}^{d_s}\big(\mathcal{X}(t,j,k)-\mu_{t,k}\big)^2.
\]
With learnable parameters $\Gamma,\mathcal{B}\in\mathbb{R}^{1\times d_s\times p}$,
\[
\mathrm{TLN}(\mathcal{X})(t,:,k)
=
\Gamma(:,:,k)\odot
\frac{\mathcal{X}(t,:,k)-\mu_{t,k}}{\sqrt{\sigma^2_{t,k}+\epsilon}}
+\mathcal{B}(:,:,k).
\]
\end{definition}

\begin{remark}
In our implementation, LayerNorm is applied in the original tensor domain slice-by-slice (mode-2 normalization within each slice). We do not rely on invariance properties under $\mathcal{L}$.
\end{remark}

\subsection{$\mathcal{L}$-Encoder layer}
\label{subsec:encoder_layer}

\begin{definition}[Tensor encoder layer]
Let $\mathcal{X}\in\mathbb{R}^{T\times d_s\times p}$. A tensor encoder layer is
\begin{align}
\mathcal{X}' &= \mathrm{TLN}\!\big(\mathcal{X}+\mathrm{MHA}_{\mathcal{L}}(\mathcal{X})\big),\\
\mathcal{Y}  &= \mathrm{TLN}\!\big(\mathcal{X}'+\mathrm{TFFN}(\mathcal{X}')\big).
\end{align}
\end{definition}

\begin{theorem}[Slice-wise equivalence of the $\mathcal{L}$-attention and $\mathcal{L}$-FFN cores]
\label{thm:core_equiv}
Let $\widehat{\mathcal{X}}=\mathcal{L}(\mathcal{X})$ and denote $\widehat{X}_i=\widehat{\mathcal{X}}(:,:,i)\in\mathbb{R}^{T\times d_s}$.
Then, for fixed parameters transformed by $\mathcal{L}$, the computations of
(i) $\mathrm{MHA}_{\mathcal{L}}(\cdot)$ and (ii) $\mathrm{TFFN}(\cdot)$ are \emph{slice-wise} in the transform domain:
for each $i\in\{1,\dots,p\}$, the corresponding output slice depends only on $\widehat{X}_i$ and the slice-$i$ transformed parameters,
and coincides with a standard attention/FFN computation at width $d_s$.
\end{theorem}

\begin{remark}[LayerNorm placement and ``exact equivalence'']
\label{rem:ln_equiv}
In our implementation, LayerNorm is applied \emph{after} mapping back to the original tensor domain (and is computed slice-by-slice along mode-2).
Because LayerNorm is nonlinear and not invariant under a general $\mathcal{L}$ transform, the \emph{full encoder layer} is not strictly separable in the transform domain.
Theorem~\ref{thm:core_equiv} therefore calibrates the exact claim to what is implemented: slice-wise equivalence holds for the attention/FFN cores,
while cross-slice coupling occurs through the alternating $\mathcal{L}^{-1}$ mixing and subsequent normalisation.
Importantly, the encoder parameter scaling and the projection/FFN FLOP reductions derived from the slice-wise cores remain valid.
\end{remark}
\subsection{$\mathcal{L}$-Decoder layer (masked self-attention + optional cross-attention)}
\label{subsec:decoder_layer}

For decoder-only LLMs (GPT-style), only masked self-attention is used. For completeness, we write the encoder--decoder form.

\begin{definition}[Tensor decoder layer]
Let $\mathcal{X}\in\mathbb{R}^{T_d\times d_s\times p}$ (decoder stream) and
$\mathcal{X}_{\text{enc}}\in\mathbb{R}^{T_e\times d_s\times p}$ (encoder stream).
\begin{align}
\mathcal{X}' &= \mathrm{TLN}\!\big(\mathcal{X}+\mathrm{MHA}_{\mathcal{L}}^{\text{masked}}(\mathcal{X})\big),\\
\mathcal{X}''&= \mathrm{TLN}\!\big(\mathcal{X}'+\mathrm{MHA}_{\mathcal{L}}^{\text{cross}}(\mathcal{X}',\mathcal{X}_{\text{enc}})\big),\\
\mathcal{Y}   &= \mathrm{TLN}\!\big(\mathcal{X}''+\mathrm{TFFN}(\mathcal{X}'')\big).
\end{align}
\end{definition}

\subsection{Complete $\mathcal{L}$-Transformer}
\label{subsec:complete_transformer}

Stacking $N$ tensor encoder and decoder layers yields a complete tensor Transformer. In tensor form:
\begin{align}
\mathcal{X}_{\text{enc}}^{(0)} &= \mathcal{X}_{\text{src}} + \mathcal{P}_{\text{enc}},\\
\mathcal{X}_{\text{enc}}^{(\ell)} &= \mathrm{TEncoderLayer}_\ell(\mathcal{X}_{\text{enc}}^{(\ell-1)}),\quad \ell=1,\dots,N,\\
\mathcal{X}_{\text{dec}}^{(0)} &= \mathcal{X}_{\text{tgt}} + \mathcal{P}_{\text{dec}},\\
\mathcal{X}_{\text{dec}}^{(\ell)} &= \mathrm{TDecoderLayer}_\ell(\mathcal{X}_{\text{dec}}^{(\ell-1)},\mathcal{X}_{\text{enc}}^{(N)}),\quad \ell=1,\dots,N.
\end{align}
Finally, unfolding back to $\mathbb{R}^{T_t\times d}$ and projecting to vocabulary size $V$ can be done via
\[
Y = \operatorname{Mat}_p\!\big(\mathcal{X}_{\text{dec}}^{(N)}\big)\,W_{\text{out}},
\qquad W_{\text{out}}\in\mathbb{R}^{d\times V}.
\]
(Alternatively one may use a tensor output projection $\mathcal{W}_{\text{out}}\in\mathbb{R}^{d_s\times V\times p}$ and then unfold.)

\subsection{Equivalence with $p$ parallel compact Transformers}
\label{subsec:equiv}

A key consequence of the $\mathcal{L}$-product definition is \emph{slice-wise decoupling} in the transform domain. Let
$\widehat{\mathcal{X}}=\mathcal{L}(\mathcal{X})$ and $\widehat{X}_i=\widehat{\mathcal{X}}(:,:,i)\in\mathbb{R}^{T\times d_s}$.
Then a tensor Transformer layer is equivalent to applying $p$ independent \emph{compact} (dimension-$d_s$) standard Transformer layers to $\widehat{X}_1,\dots,\widehat{X}_p$ with transformed parameters, stacking the results, and applying $\mathcal{L}^{-1}$.

Formally, if ${T}_{\mathrm{std}}^{(i)}:\mathbb{R}^{T\times d_s}\to\mathbb{R}^{T\times d_s}$ denotes the slice-$i$ standard Transformer map (with slice-specific transformed weights), then the tensor output satisfies
\begin{equation}
\label{eq:equiv}
\mathcal{Z}
=
\mathcal{L}^{-1}\!\Big(
\text{Stack}_3\big(
{T}_{\mathrm{std}}^{(1)}(\widehat{X}_1),\dots,{T}_{\mathrm{std}}^{(p)}(\widehat{X}_p)
\big)
\Big).
\end{equation}
\paragraph{Coupling across slices.}
Although the computation is slice-wise in the transform domain, the model is not equivalent to a static partition of the embedding dimension.
After each block, applying the inverse transform $\mathcal{L}^{-1}$ mixes the $p$ spectral channels, so the input slices of the next layer are linear combinations of the previous layer’s slice outputs.
As a result, information can propagate across slices from layer to layer through the alternating $\mathcal{L}$ / $\mathcal{L}^{-1}$ mappings, yielding structured cross-slice interactions even though each block is computed slice-wise in the $\mathcal{L}$-domain.
\begin{corollary}[Parallel computation]
\label{cor:parallel}
The $p$ slice-wise Transformers are independent in the transform domain and can be executed in parallel.
\end{corollary}

\subsection{Training and gradients through the $\mathcal{L}$-product}
\label{subsec:training}

Differentiability is preserved because $\mathcal{L}$ and $\mathcal{L}^{-1}$ are linear maps and the facewise multiplications are standard matrix products in the transform domain.

\begin{proposition}[Backpropagation through $\mathcal{L}$-products]
\label{prop:grad_Lprod}
Let $\mathcal{Y}=\mathcal{X}*_{\mathcal{L}}\mathcal{W}$ with
$\mathcal{X}\in\mathbb{R}^{T\times d_s\times p}$ and
$\mathcal{W}\in\mathbb{R}^{d_s\times d'\times p}$.
Given $G=\frac{\partial \mathcal{J}}{\partial \mathcal{Y}}$, the gradients are
\[
\frac{\partial \mathcal{J}}{\partial \mathcal{X}} = G*_{\mathcal{L}}\mathcal{W}^{T},
\qquad
\frac{\partial \mathcal{J}}{\partial \mathcal{W}} = \mathcal{X}^{T}*_{\mathcal{L}}G,
\]
where $(\cdot)^T$ is the $\mathcal{L}$-transpose.
\end{proposition}

\subsection{Computational complexity}
\label{subsec:complexity}

We compare per-layer FLOPs for a standard Transformer operating on $d$ against a tensor Transformer with factor $p$ and slice width $d_s=d/p$.
Let $d_{ff}=4d$ and $d_{ff,s}=4d_s$.

\begin{table}[h!]
\centering
\caption{Per-layer FLOPs comparison. Projection and FFN costs drop by a factor $\approx p$, while attention score computation remains unchanged (it scales as $T^2$). }
\label{tab:complexity}
\setlength{\tabcolsep}{6pt}
\renewcommand{\arraystretch}{1.15}
\small
\begin{tabularx}{\textwidth}{@{}>{\RaggedRight\arraybackslash}p{4.4cm}
                        >{\RaggedRight\arraybackslash}X
                        >{\RaggedRight\arraybackslash}X@{}}
\toprule
\textbf{Operation} & \textbf{Standard} & \textbf{Tensor ($\mathcal{L}$-product)} \\
\midrule
$Q,K,V$ projections
& $3T d^2$
& $3T\,p\,d_s^2 \;=\; 3T d^2/p$ \\

Attention scores ($QK^\top$)
& $T^2 d$
& $p\,T^2 d_s \;=\; T^2 d$ \emph{(unchanged)} \\

Attention $\times V$
& $T^2 d$
& $T^2 d$ \emph{(unchanged)} \\

Output projection
& $T d^2$
& $T d^2/p$ \\

FFN (two layers)
& $8T d^2$
& $8T d^2/p$ \\

$\mathcal{L}$-transform (DCT/IDCT)
& ---
& $T d \log p$ \\

\midrule
\textbf{Total}
& $12T d^2 + 2T^2 d$
& $12T d^2/p + 2T^2 d + T d\log p$ \\
\bottomrule
\end{tabularx}
\end{table}
\paragraph{FLOPs vs.\ wall-clock time.}
Table~\ref{tab:complexity} accounts for arithmetic cost, but measured runtime is also influenced by
(i) extra memory-bound passes introduced by the DCT/IDCT steps, and
(ii) kernel-launch/parallelism effects when the $p$ slice-wise sub-encoders are executed sequentially.

With the batched parallel implementation noted after Algorithm~\ref{alg:mha}, effect (ii) is largely removed,
and the remaining overhead is dominated by the transform steps and framework-level memory traffic.

The standard Transformer has per-layer cost $\mathcal{O}(T d^2 + T^2 d)$.
The tensor Transformer reduces the \emph{projection} and \emph{FFN} terms to $\mathcal{O}(T d^2/p)$, while the attention terms remain $\mathcal{O}(T^2 d)$.
The transform overhead $\mathcal{O}(T d\log p)$ is typically small.

\begin{remark}[When the speedup is largest]
When $d \gg T$, the $T d^2$ terms dominate and the tensor Transformer yields an effective $\approx 1/p$ reduction in per-layer FLOPs.
When $T \gg d$, the attention term $T^2 d$ dominates and the speedup diminishes; the tensor framework can then be combined with efficient attention approximations \cite{kitaev2020reformer,wang2020linformer,choromanski2020rethinking,dao2022flashattention}.
\end{remark}

\begin{remark}[Memory]
Parameter storage drops by $\approx 1/p$ (the dominant $d^2$ terms), while attention-map storage (order $T^2$) is unchanged.
\end{remark}

\section{Numerical experiments}
\label{sec:experiments}
We evaluate the proposed tensor $\mathcal{L}$-product Transformer on text classification, comparing it to a standard Transformer encoder (Std) under controlled training and preprocessing.
We study two axes: (i) \emph{spectral weighting} (PE) at fixed $p{=}4$, and (ii) \emph{tensor factorization} through the decomposition factor $p$ (with an ablation at $p\in\{2,4\}$ on IMDB).
Results are reported as mean $\pm$ standard deviation over three independent seeds (42, 123, 7).
We consider two benchmarks: IMDB (binary sentiment, 25\,K/25\,K train/test)~\cite{maas2011learning} and AG~News (4-class topic classification, 120\,K/7.6\,K)~\cite{zhang2015character}.
We report results at three widths $d\in\{128,256,768\}$; the largest setting ($d{=}768$) matches the embedding dimension of BERT-base \cite{devlin2019bert} but is trained \emph{from scratch}\footnote{We use “BERT-base width” only to indicate the embedding dimension $d=768$. We do not initialize from pretrained BERT weights; all models here are trained from random initialization under the same training recipe.}.

\medskip

\noindent\textbf{Notation.}
\textbf{Std} denotes the vanilla Transformer encoder baseline.
\textbf{T$N$-\textit{pe}} denotes a tensor model with factor $p{=}N$ and spectral weighting \textit{pe}.
At $p{=}4$, we consider: T4-standard ($\alpha_k{=}1$), T4-linear ($\alpha_k{=}k/p$), T4-harmonic ($\alpha_k{=}k$),
T4-exponential ($\alpha_k{=}2^{(k-1)/(p-1)}$), and T4-learnable (learned $\alpha_k$, adding $p$ trainable scalars with the same weight decay as all other parameters).

\medskip

\noindent\textbf{Empirical message.}
The central question is whether tensorizing the representation space yields an encoder parameter reduction close to $1/p$ while preserving accuracy.
At moderate width, we observe a clear pattern: strong gains on IMDB and a modest accuracy cost on AG~News at $d{=}256$.
At larger width ($d{=}768$), the gap closes: the tensor model matches within run-to-run variability (3 seeds) with Std while compressing the encoder by $4\times$ and reducing peak memory by 15\%.
This scaling trend is the key evidence supporting the relevance of the method beyond small models.

\subsection{Experimental setup and evaluation protocol}
\label{sec:setup}

\paragraph{Architecture.}
All configurations use $n_{\text{layers}}{=}4$ encoder layers, a vocabulary of 30\,000 (byte-pair encoding), and maximum sequence length $T{=}128$, with identical truncation and padding across models.
Table~\ref{tab:arch_shared} summarizes the shared settings, and Table~\ref{tab:arch_by_task} reports task-specific widths.
\begin{table}[h!]
\centering
\caption{Shared architecture settings across all experiments. Tokenization uses byte-pair encoding (BPE) \cite{sennrich2016neural}.}
\label{tab:arch_shared}
\small
\setlength{\tabcolsep}{7pt}
\renewcommand{\arraystretch}{1.15}
\begin{tabularx}{0.92\textwidth}{@{}>{\RaggedRight\arraybackslash}p{4.2cm}
                                >{\RaggedRight\arraybackslash}X@{}}
\toprule
\textbf{Component} & \textbf{Value} \\
\midrule
Encoder depth & $n_{\text{layers}} = 4$ \\
Tokenizer / vocabulary & BPE, 30{,}000 types \\
Max sequence length & $T = 128$ (identical truncation/padding) \\
\bottomrule
\end{tabularx}
\end{table}
\begin{table}[h!]
\centering
\caption{Task-specific architecture settings.}
\label{tab:arch_by_task}
\small
\setlength{\tabcolsep}{8pt}
\renewcommand{\arraystretch}{1.15}
\begin{tabular}{@{}lccc@{}}
\toprule
\textbf{Setting} & \textbf{$d$} & \textbf{$H$} & \textbf{$d_{\mathrm{ff}}$} \\
\midrule
IMDB & 128 & 4 & 512 \\
AG~News (moderate) & 256 & 4 & 1024 \\
AG~News (BERT-width) & 768 & 8 & 3072 \\
\bottomrule
\end{tabular}
\end{table}
For a tensor model with factor $p$, the embedding output $X\in\mathbb{R}^{T\times d}$ is reshaped into
$\mathcal{X}\in\mathbb{R}^{T\times d_s\times p}$ with $d_s=d/p$, processed slice-wise in the DCT domain \cite{Ahmed1974DCT}, and reassembled via the inverse $\mathcal{L}$-transform.

\paragraph{Training.}
Table~\ref{tab:train_shared} reports the optimizer and system configuration shared by all experiments, and Table~\ref{tab:train_by_task} gives dataset-specific schedules.
\begin{table}[h!]
\centering
\caption{Shared training configuration across all experiments.}
\label{tab:train_shared}
\small
\setlength{\tabcolsep}{7pt}
\renewcommand{\arraystretch}{1.15}
\begin{tabularx}{\textwidth}{@{}>{\bfseries}p{3.6cm} >{\RaggedRight\arraybackslash}X@{}}
\toprule
Component & Value \\
\midrule
Optimizer &
AdamW (lr $=3\times 10^{-4}$, weight decay $=0.01$) \\
Scheduler &
OneCycleLR (10\% linear warmup, cosine annealing to $10^{-5}$) \cite{smith2017cyclical,smith2019super} \\
Gradient clipping &
$1.0$ \\
Precision &
Mixed precision (PyTorch 2.1, \texttt{torch.cuda.amp}) \cite{micikevicius2017mixed} \\
Hardware &
NVIDIA Tesla T4 (16\,GB), CUDA 12.1 \\
Reproducibility &
\texttt{torch.manual\_seed}, \texttt{numpy.random.seed}, \texttt{random.seed};
\texttt{cudnn.deterministic=True} \\
\bottomrule
\end{tabularx}
\end{table}
\begin{table}[h!]
\centering
\caption{Dataset-specific training schedules.}
\label{tab:train_by_task}
\small
\setlength{\tabcolsep}{8pt}
\renewcommand{\arraystretch}{1.15}
\begin{tabular}{@{}lcc@{}}
\toprule
\textbf{Setting} & \textbf{Epochs} & \textbf{Batch size} \\
\midrule
IMDB ($d=128$) & 20 & 128 \\
AG~News ($d=256$) & 5 & 128 \\
AG~News ($d=768$) & 5 & 64 \\
\bottomrule
\end{tabular}
\end{table}

\paragraph{Evaluation.}
Each configuration is run for all 3 seeds.
Each model is trained for a fixed number of epochs (20 for IMDB, 5 for AG~News) with no early stopping and no checkpoint selection.
We report test accuracy at the \textbf{final epoch}, computed on the held-out test set (25\,000 samples for IMDB; 7\,600 for AG~News). All models plateau well before the final epoch (Appendix~\ref{app:learning_curves}), and the last-epoch accuracy is stable across seeds. Peak GPU memory is measured via \texttt{torch.cuda.max\_memory\_allocated}.
Inference latency is the mean of 3 forward passes on the full test set under \texttt{torch.no\_grad()} with mixed precision. Inference timing is measured after training and does not affect any training or selection decision

\subsection{Fair comparison protocol}
\label{sec:fair}

All models share identical embedding and classifier layers; only the encoder varies.
This isolates the encoder as the sole source of architectural and parameter-count differences.
Table~\ref{tab:model_configs} summarizes what is fixed and what varies.

\begin{table}[h!]
\centering
\caption{Controlled comparison: fixed versus varying components.}
\label{tab:model_configs}
\small
\setlength{\tabcolsep}{7pt}
\renewcommand{\arraystretch}{1.15}
\begin{tabularx}{\textwidth}{@{}>{\RaggedRight\arraybackslash}X >{\RaggedRight\arraybackslash}X@{}}
\toprule
\textbf{Fixed across all models} & \textbf{Varies (encoder only)} \\
\midrule
Embedding layer (vocabulary $\times d$) &
Encoder type (Std vs.\ $\mathcal{L}$-product) \\

Classifier head (mean-pool + linear) &
Decomposition factor $p$ \\

Training recipe (optimizer, scheduler, hyperparameters) &
Spectral weighting strategy (\textit{pe}) \\

Data protocol (sequence length $T$, tokenizer, padding/truncation, seeds) &
Encoder parameter count (and resulting compression ratio) \\
\bottomrule
\end{tabularx}
\end{table}

For a tensor model with factor $p$, the encoder operates on $p$ sub-encoders of width $d_s=d/p$, yielding approximately $1/p$ the encoder parameters of Std.
At small $d$, the embedding layer dominates total parameters (e.g.\ IMDB), so encoder compression is not fully reflected as total-model compression.
At larger $d$, the encoder becomes a dominant fraction of parameters, so encoder compression translates into substantial end-to-end savings (Section~\ref{sec:scaling}).

\paragraph{Parameter-matched baseline.}
To disentangle architectural effects from pure parameter reduction, we include a parameter-matched (PM) standard baseline that approximately matches the encoder parameter count of T4 ($p{=}4$).
Specifically, \textbf{Std-1L} is a single-layer standard encoder at full width $d$.
At $d{=}128$ this yields $247{,}808$ encoder parameters ($0.250\times$ Std), closely matching T4-linear's $254{,}016$ ($0.256\times$).
At $d{=}256$ it yields $987{,}136$ encoder parameters ($0.250\times$ Std), matching T4-linear's $999{,}552$ ($0.253\times$).
This comparison isolates the architectural question: at equal encoder budget, is it better to invest in a single full-width layer or in four tensor-factored layers.
Table~\ref{tab:pm_summary} shows that T4-linear consistently outperforms Std-1L on both datasets, indicating that the tensor $\mathcal{L}$-product structure provides more than a simple capacity reduction.

\begin{table}[h!]
  \centering
  \caption{Parameter-matched comparison (mean $\pm$ std over 3 seeds). Encoder parameters and ratio relative to Std.}
  \label{tab:pm_summary}
  \small
  \begin{tabular}{llccc}
    \toprule
    \textbf{Dataset} & \textbf{Model} & \textbf{Acc.\ (\%)} & \textbf{Enc.\ params} & \textbf{Ratio} \\
    \midrule
    \multirow{3}{*}{IMDB ($d{=}128$)}
      & Std (4L, $d{=}128$)        & $80.77 \pm 0.24$ & 991\,K  & $1.000\times$ \\
      & Std-1L (1L, $d{=}128$)     & $77.50 \pm 0.28$ & 248\,K  & $0.250\times$ \\
      & T4-linear (4L, $d_s{=}32$) & $\mathbf{81.65 \pm 0.06}$ & 254\,K  & $0.256\times$ \\
    \midrule
    \multirow{3}{*}{AG~News ($d{=}256$)}
      & Std (4L, $d{=}256$)        & $\mathbf{91.40 \pm 0.31}$ & 3\,949\,K & $1.000\times$ \\
      & Std-1L (1L, $d{=}256$)     & $86.97 \pm 0.15$ & 987\,K    & $0.250\times$ \\
      & T4-linear (4L, $d_s{=}64$) & $90.76 \pm 0.13$ & 1\,000\,K & $0.253\times$ \\
    \bottomrule
  \end{tabular}
\end{table}

\subsection{Main results and PE strategy comparison}
\label{sec:main_results}

\begin{figure}[h!]
  \centering
  \includegraphics[width=\textwidth]{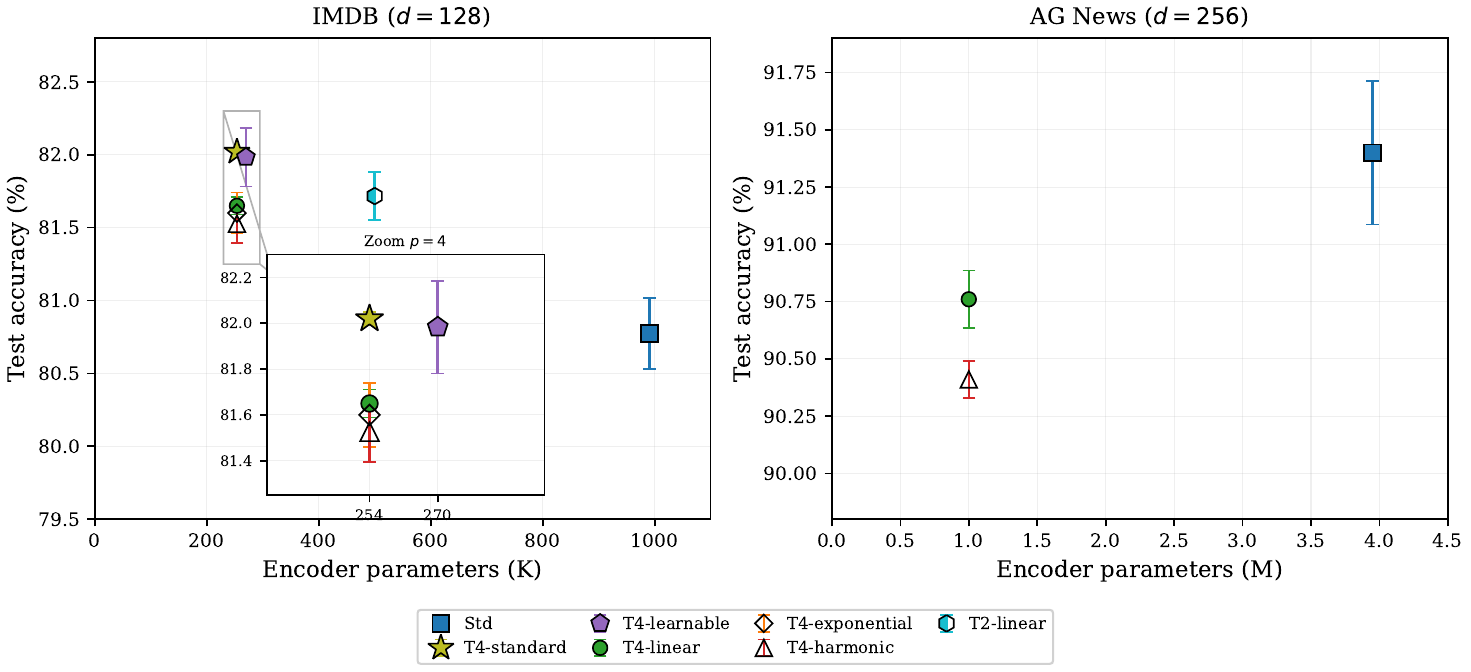}
  \caption{Encoder parameters versus test accuracy on IMDB (left) and AG~News (right). Error bars show $\pm 1$ std over 3 seeds. Tensor models achieve competitive or superior accuracy while using approximately $4\times$ fewer encoder parameters at $p=4$.}
  \label{fig:params_vs_acc}
\end{figure}

Tables~\ref{tab:main_imdb} and~\ref{tab:main_agnews} report test accuracy, encoder parameters, and compression ratios on IMDB and AG~News ($d{=}256$).
Figure~\ref{fig:params_vs_acc} summarizes the encoder parameter--accuracy trade-off.

\begin{table}[h!]
  \centering
  \caption{IMDB results ($d{=}128$, 20 epochs, 3 seeds). Embedding (3.84\,M) and classifier are shared. Best in \textbf{bold}.}
  \label{tab:main_imdb}
  \small
  \begin{tabular}{lcccc}
    \toprule
    \textbf{Model} & $p$ & \textbf{Acc.\ (\%)} & \textbf{Encoder} & \textbf{Enc.\ Ratio} \\
    \midrule
    Std            & --- & $80.77 \pm 0.24$ & 991\,K & $1.000\times$ \\
    \midrule
    T4-standard    & 4 & $\mathbf{82.02 \pm 0.03}$ & 254\,K & $0.256\times$ \\
    T4-learnable   & 4 & $81.98 \pm 0.20$ & 270\,K & $0.273\times$ \\
    T4-linear      & 4 & $81.65 \pm 0.06$ & 254\,K & $0.256\times$ \\
    T4-exponential & 4 & $81.60 \pm 0.14$ & 254\,K & $0.256\times$ \\
    T4-harmonic    & 4 & $81.53 \pm 0.13$ & 254\,K & $0.256\times$ \\
    \midrule
    T2-linear      & 2 & $81.72 \pm 0.17$ & 500\,K & $0.504\times$ \\
    \bottomrule
  \end{tabular}
\end{table}

On IMDB, encoder compression does not degrade accuracy; instead, all $p{=}4$ tensor variants improve upon Std.
The best configuration (T4-standard) reaches $82.02\% \pm 0.03$ versus $80.77\% \pm 0.24$ for Std (+1.25\,pp), while using $0.256\times$ encoder parameters.
The narrow spread across PE strategies (0.49\,pp from best to worst) indicates that performance is driven primarily by the tensor factorization itself, with spectral weighting acting as a secondary tuning knob.
The parameter-matched comparison in Table~\ref{tab:pm_summary} strengthens this point: at the same encoder budget, T4-linear substantially outperforms Std-1L, showing that the gain is not explained by mere parameter reduction.

\begin{table}[h!]
  \centering
  \caption{AG~News results ($d{=}256$, 5 epochs, 3 seeds). Best in \textbf{bold}.}
  \label{tab:main_agnews}
  \small
  \begin{tabular}{lcccc}
    \toprule
    \textbf{Model} & $p$ & \textbf{Acc.\ (\%)} & \textbf{Encoder} & \textbf{Enc.\ Ratio} \\
    \midrule
    Std         & --- & $\mathbf{91.40 \pm 0.31}$ & 3.95\,M & $1.000\times$ \\
    \midrule
    T4-linear   & 4 & $90.76 \pm 0.13$ & 1.00\,M & $0.253\times$ \\
    T4-harmonic & 4 & $90.41 \pm 0.08$ & 1.00\,M & $0.253\times$ \\
    \bottomrule
  \end{tabular}
\end{table}

On AG~News at moderate width, the tensorized encoder yields a clear compression--accuracy trade-off.
T4-linear compresses the encoder by $4\times$ (3.95\,M $\to$ 1.00\,M) at a modest cost of 0.64\,pp relative to Std.
This result is best interpreted together with the scaling experiment in Section~\ref{sec:scaling}, where the gap closes at $d{=}768$.
The parameter-matched baseline in Table~\ref{tab:pm_summary} further clarifies the story: at the same encoder budget, T4-linear substantially outperforms Std-1L, indicating that the tensor structure is a more effective use of parameters than simply shrinking depth.

Figure~\ref{fig:params_vs_acc} visualizes the key empirical trend: tensor factorization moves models leftward (fewer encoder parameters) with limited or no loss in accuracy, and on IMDB with a clear improvement.
On AG~News at $d{=}256$, tensor points lie slightly below Std in accuracy but far to the left in encoder size, reflecting a strong compression benefit.

\subsection{Spectral weighting (PE) ablation}
\label{sec:pe_ablation}

\begin{figure}[h!]
  \centering
  \includegraphics[width=0.7\columnwidth]{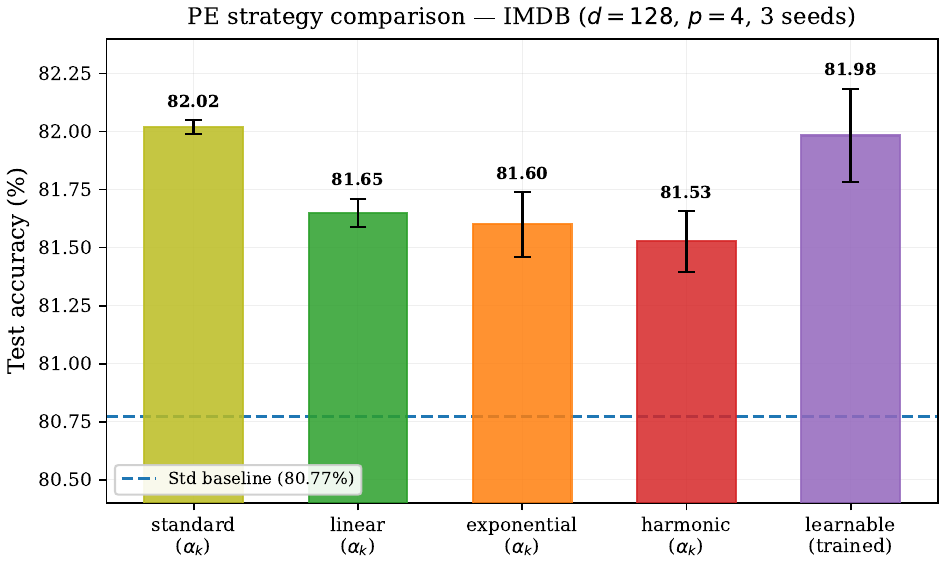}
  \caption{PE strategy comparison on IMDB at $p=4$ (mean $\pm$ std, 3 seeds). The dashed line shows the full-width Std baseline. All strategies outperform Std, and the total spread across strategies is 0.49\,pp.}
  \label{fig:pe_comparison}
\end{figure}

Figure~\ref{fig:pe_comparison} compares all five PE strategies at $p{=}4$ on IMDB.
The full sweep (including AG~News) is reported in Appendix~\ref{app:pe_sweep}.

Two conclusions emerge.
First, spectral weighting is not brittle: all strategies remain competitive, and the performance spread is small.
Second, the best weighting is dataset-dependent: identity weighting (T4-standard) is best on IMDB, while linear weighting is best on AG~News at $d{=}256$ (Appendix~\ref{app:pe_sweep}).
This supports the interpretation that $\alpha_k$ induces a frequency-domain bias whose optimal form depends on the task.

\subsection{Scaling to BERT-base dimension}
\label{sec:scaling}

To test the method at larger width, we train from scratch on AG~News at $d{=}768$ with $H{=}8$, $d_{\mathrm{ff}}{=}3072$, and $p{=}4$ (so $d_s{=}192$).
At this width, the encoder dominates the parameter budget, and $1/p$ encoder compression translates into substantial end-to-end savings.

\begin{table}[h!]
  \centering
  \caption{AG~News at $d=768$ (BERT-base width), $p=4$, from scratch, 3 seeds. The tensor model compresses the encoder by $4\times$ and reduces peak memory while maintaining accuracy.}
  \label{tab:bert_scale}
  \small
  \begin{tabular}{lcccc}
    \toprule
    \textbf{Model} & \textbf{Acc.\ (\%)} & \textbf{Encoder} & \textbf{Ratio} & \textbf{Mem} \\
    \midrule
    Std       & $91.47 \pm 0.09$ & 28.4\,M & $1.00\times$ & 1.95\,GB \\
    T4-linear & $\mathbf{91.52 \pm 0.21}$ & 7.1\,M & $0.25\times$ & 1.66\,GB \\
    \bottomrule
  \end{tabular}
\end{table}

\begin{table}[h!]
  \centering
  \caption{Parameter breakdown at $d=768$. Encoder compression becomes a model-level reduction because the encoder accounts for the majority of parameters.}
  \label{tab:param_breakdown_768}
  \small
  \begin{tabular}{lcc}
    \toprule
    \textbf{Component} & \textbf{Std} & \textbf{T4-linear} \\
    \midrule
    Embedding   & 23.0\,M (45\%) & 23.0\,M (76\%) \\
    Encoder     & 28.4\,M (55\%) & 7.1\,M (24\%) \\
    \midrule
    Total       & 51.4\,M        & 30.2\,M ($0.59\times$) \\
    \bottomrule
  \end{tabular}
\end{table}

At BERT-base width, T4-linear matches within run-to-run variability (3 seeds) with Std while compressing the encoder from 28.4\,M to 7.1\,M parameters (Table~\ref{tab:bert_scale}).
Unlike moderate widths, encoder compression is no longer masked by the embedding layer: the total model drops from 51.4\,M to 30.2\,M parameters (Table~\ref{tab:param_breakdown_768}) and peak memory falls by 15\%.
This is the regime where structural compression matters most in practice.

\begin{figure}[h!]
  \centering
  \includegraphics[width=0.65\columnwidth]{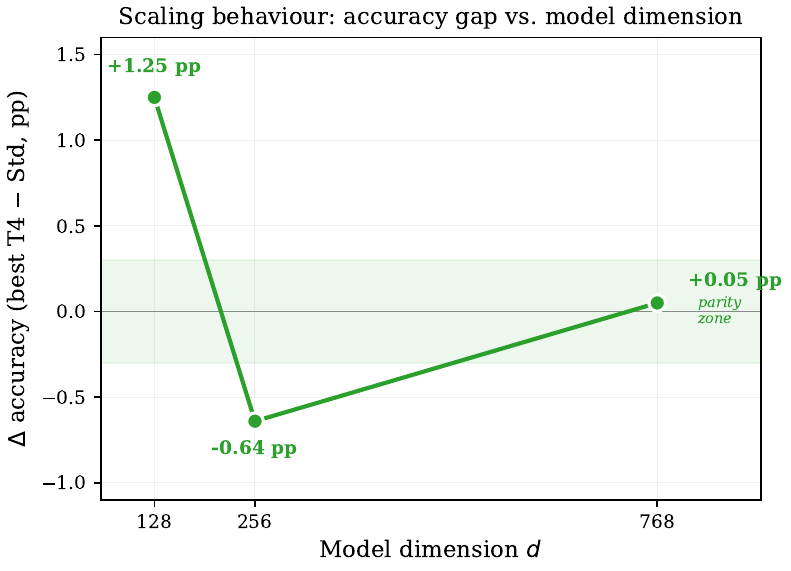}
  \caption{Accuracy gap (best T4 minus Std) as a function of model width $d$. The shaded band indicates a $\pm 0.3$\,pp parity region. At $d=768$, the tensor model reaches parity while compressing the encoder by $4\times$.}
  \label{fig:scaling}
\end{figure}

The scaling trend suggests that per-slice width $d_s=d/p$ is a key driver.
At $d{=}256$, $d_s{=}64$ may be slightly restrictive for AG~News; at $d{=}768$, $d_s{=}192$ appears sufficient to match Std performance while retaining the $4\times$ encoder compression.
This aligns with the theory: in the transform domain, each slice behaves as a compact Transformer whose capacity depends on $d_s$.

\subsection{Efficiency analysis}
\label{sec:efficiency}

Table~\ref{tab:efficiency} reports wall-clock and memory measurements that complement the FLOPs analysis in Section~5.14.
All measurements are taken on a single Tesla T4 GPU with mixed precision and sequence length $T{=}128$.
At moderate widths ($d\le 256$), sequential slice execution introduces wall-clock overhead despite the $4\times$ encoder reduction.
At $d{=}768$, the compute reduction dominates: epoch time improves and peak memory decreases substantially.

\begin{table}[h!]
  \centering
  \caption{Efficiency comparison at equal depth (4 layers, mean over 3 seeds). Train throughput is measured per epoch; inference uses the full test set under \texttt{torch.no\_grad()} with mixed precision.}
  \label{tab:efficiency}
  \small
  \resizebox{\textwidth}{!}{
  \begin{tabular}{llcccccc}
    \toprule
    \textbf{Config} & \textbf{Model}
      & \textbf{Enc.\ params}
      & \textbf{Ratio}
      & \textbf{Train tok/s}
      & \textbf{Epoch (s)}
      & \textbf{Peak mem}
      & \textbf{Batch} \\
    \midrule
    \multirow{2}{*}{IMDB $d{=}128$}
      & Std        & 991\,K   & $1.00\times$ & 123\,K & 26\,s  & 0.50\,GB & 128 / 256 \\
      & T4-linear  & 254\,K   & $0.26\times$ &  94\,K & 34\,s  & 0.51\,GB & 128 / 256 \\
    \midrule
    \multirow{2}{*}{AG~News $d{=}256$}
      & Std        & 3\,949\,K & $1.00\times$ & 177\,K & 87\,s  & 0.99\,GB & 128 / 256 \\
      & T4-linear  & 1\,000\,K & $0.25\times$ & 132\,K & 116\,s & 0.96\,GB & 128 / 256 \\
    \midrule
    \multirow{2}{*}{AG~News $d{=}768$}
      & Std        & 28\,352\,K & $1.00\times$ & 60\,K & 257\,s & 1.95\,GB & 64 / 128 \\
      & T4-linear  &  7\,118\,K & $0.25\times$ & 64\,K & 241\,s & \textbf{1.66\,GB} & 64 / 128 \\
    \bottomrule
  \end{tabular}}
  \smallskip
  \parbox{\textwidth}{\footnotesize
  Batch column: train / inference batch sizes.
  At $d\le 256$, sequential slice execution introduces $\approx 30\%$ wall-clock overhead despite the $4\times$ encoder reduction.
  At $d=768$, the compute reduction dominates: T4-linear is 6\% faster per epoch and uses 15\% less peak memory.}
\end{table}

\subsection{Summary of findings}
\label{sec:findings}

Table~\ref{tab:summary} consolidates the main results across the three widths.

\begin{table}[h!]
  \centering
  \caption{Summary across all configurations (mean $\pm$ std, 3 seeds). $\Delta$ is the accuracy difference of the best tensor variant versus Std.}
  \label{tab:summary}
  \small
  \setlength{\tabcolsep}{3.5pt}
  \begin{tabular}{lcccc}
    \toprule
    \textbf{Configuration} & \textbf{Std} & \textbf{Best T4} & $\boldsymbol{\Delta}$ & \textbf{Enc.\ Saving} \\
    \midrule
    IMDB $d=128$ & $80.77_{\pm 0.24}$ & $82.02_{\pm 0.03}$ (std) & $+1.25$ & 991\,K $\to$ 254\,K \\
    AG~News $d=256$ & $91.40_{\pm 0.31}$ & $90.76_{\pm 0.13}$ (lin) & $-0.64$ & 3.95\,M $\to$ 1.00\,M \\
    AG~News $d=768$ & $91.47_{\pm 0.09}$ & $91.52_{\pm 0.21}$ (lin) & $+0.05$ & 28.4\,M $\to$ 7.1\,M \\
    \bottomrule
  \end{tabular}
\end{table}

Across widths, the tensor $\mathcal{L}$-product encoder achieves approximately $4\times$ encoder compression with competitive accuracy.
On IMDB, it improves performance by 1.25\,pp while substantially reducing variance.
On AG~News at $d=256$, it trades 0.64\,pp in accuracy for a $4\times$ reduction in encoder parameters; the parameter-matched baseline confirms this is still a strong use of parameters.
At $d=768$, the method reaches parity while delivering a 41\% total-model reduction (Table~\ref{tab:param_breakdown_768}) and a 15\% peak-memory reduction (Table~\ref{tab:bert_scale}).
Taken together, these results support the main claim: tensor factorization becomes increasingly effective as width grows, precisely the regime where encoder compression is most valuable.
\section{Conclusion}
\label{sec:conclusion}

We introduced a structured multidimensional representation learning framework for Transformers based on the
$\mathcal{L}$-product for third-order tensors. By reshaping token embeddings from $X\in\mathbb{R}^{T\times d}$
into $\mathcal{X}\in\mathbb{R}^{T\times d_s\times p}$ with $d_s=d/p$, and applying attention and feed-forward
operations slice-wise in the transform domain, the resulting encoder becomes spectrally equivalent to $p$ parallel
compact Transformers operating on width $d_s$. This yields an $\approx 1/p$ reduction of encoder parameters
(up to lower-order terms such as biases and normalization parameters), while preserving standard Transformer
semantics after inverse transformation.

Empirically, on IMDB the tensorized encoder achieves improved accuracy under a $4\times$ encoder compression,
and the parameter-matched baseline confirms that the gains are not explained by parameter count alone.
On AG~News at moderate width ($d=256$), the method exhibits a clear compression--accuracy trade-off, while at
BERT-base width ($d=768$) the tensor model reaches statistical parity with the baseline while compressing the
encoder by $4\times$ and reducing peak GPU memory.

The approach does not reduce the quadratic attention-map term (both compute and memory scale with $T^2$),
so wall-clock gains depend on the regime and implementation. In particular, a sequential slice execution can
introduce overhead at moderate widths, and the method currently assumes $p\mid d$ for a clean tensorization.

Several directions follow naturally: (i) combining the tensor $\mathcal{L}$-product encoder with efficient attention
approximations to address the $T^2$ bottleneck; (ii) implementing fully batched slice execution to better translate
theoretical FLOP reductions into runtime improvements; (iii) extending evaluation to broader benchmarks and tasks,
including pretrained fine-tuning settings; and (iv) exploring alternative orthogonal transforms and learned
transform operators to further adapt the spectral inductive bias to different domains.
\section{Declaration of Generative AI and AI-assisted technologies in the writing process}
 During the preparation of this work, the authors used ChatGPT to improve \LaTeX{} tables.
After using this tool, the author reviewed and edited the content as needed and take
full responsibility for the publication's content.
\section{ Declaration of Competing Interest}
We hereby declare that we do not have any financial and personal relationships with other people
or organizations that could inappropriately influence (bias) our work.
\clearpage
\appendix
\section{Additional theoretical and implementation details}
\subsection{Tensor unfolding conventions}
\label{app:unfolding-conv}
We follow standard tensor notation and unfolding conventions as in
\cite{KoldaBader2009}; all derivations rely only on the standard
identity for the $n$-mode product:
\[
(\mathcal{A}\times_n X)_{(n)} = X\,\mathcal{A}_{(n)}.
\]
Unfolding  converts a tensor into a matrix by
choosing one mode to index the rows and concatenating all remaining modes into the columns.
This operation is useful because many multilinear identities (notably those involving the $n$-mode product)
can be expressed compactly through standard matrix multiplication.

Let $\mathcal{A}\in\mathbb{R}^{n_1\times\cdots\times n_N}$. The \emph{mode-$n$ unfolding}
$\mathcal{A}_{(n)}$ is the matrix in
\[
\mathcal{A}_{(n)} \in \mathbb{R}^{n_n \times \prod_{k\ne n} n_k},
\]
obtained by fixing the $n$th index and enumerating all remaining indices in a consistent order.

\medskip
\noindent\textbf{Third-order case.}
For $\mathcal{A}\in\mathbb{R}^{n_1\times n_2\times n_3}$, the unfoldings satisfy
\[
\mathcal{A}_{(1)}\in\mathbb{R}^{n_1\times (n_2 n_3)},\qquad
\mathcal{A}_{(2)}\in\mathbb{R}^{n_2\times (n_1 n_3)},\qquad
\mathcal{A}_{(3)}\in\mathbb{R}^{n_3\times (n_1 n_2)}.
\]
In this paper, the mode-$3$ unfolding is particularly important because the $\mathcal{L}$-transform
acts along the tube dimension (mode~3). Concretely, writing $\mathcal{A}_{(3)}$ stacks all tubes
$\mathcal{A}(i,j,:)\in\mathbb{R}^{n_3}$ as columns.

When we tensorize token embeddings into $\mathcal{X}\in\mathbb{R}^{T\times d_s\times p}$, the operator
$\mathcal{L}(\mathcal{X})=\mathcal{X}\times_3 Z$ can be viewed in unfolded form as
\[
\big(\mathcal{L}(\mathcal{X})\big)_{(3)} = Z\,\mathcal{X}_{(3)},
\]
i.e., a standard matrix multiplication on the tube-stacked representation. This perspective clarifies why
the transform is linear, differentiable, and efficiently implementable.
\subsection{Detailed $\mathcal{L}$-Multi-Head Attention}
\label{app:mha_detail}

This appendix expands Algorithm~\ref{alg:mha} into a fully explicit, implementation-level specification of
$\mathcal{L}$-Multi-Head Attention. The goal is twofold.

\paragraph{(i) Reproducibility and shape bookkeeping.}
We enumerate every intermediate tensor/matrix and its dimensions \emph{per slice} ($i=1,\dots,p$) and \emph{per head}
($j=1,\dots,h$), including the slice-specific projected queries/keys/values, attention score matrices, and the final
per-slice output projection. This resolves common sources of ambiguity in tensorized attention implementations
(e.g., where the head dimension is split, which axis is treated as the batch axis, and how concatenation is performed).

\paragraph{(ii) Operational meaning of the slice-wise equivalence.}
By writing the computation slice-by-slice in the $\mathcal{L}$-domain, the appendix makes the equivalence statement
concrete: for each fixed slice $i$, the attention update is \emph{exactly} the standard scaled dot-product multi-head
attention applied to $\widehat{X}^{(i)}\in\mathbb{R}^{T\times d_s}$ with slice-specific transformed parameters.
This clarifies what the method changes (projection/FFN parameterization and compute scaling with $d_s=d/p$) and what it
does not change (the $T\times T$ attention-map structure and the quadratic-in-$T$ score computation).

\paragraph{Parallelization note.}
Although presented with an explicit slice loop for clarity, the detailed formulation also exposes the natural batching
strategy used in practice: treat the slice index (and optionally the head index) as a batch dimension and implement the
slice-wise multiplications via batched matrix multiplications / fused attention kernels. This is the key step that
allows the theoretical per-slice independence in the transform domain to translate into efficient GPU execution.
\begin{algorithm}[H] \caption{$\mathcal{L}$-Multi-Head Attention (detailed spectral implementation)} \label{alg:mha_detailed} \begin{algorithmic}[1] \REQUIRE Input $\mathcal{X}\in\mathbb{R}^{T\times d_s\times p}$; positional encoding $\mathcal{P}\in\mathbb{R}^{T\times d_s\times p}$; heads $h$, $d_s=h\,d_h$; weights $\{\mathcal{W}_Q^{(j)},\mathcal{W}_K^{(j)},\mathcal{W}_V^{(j)}\}_{j=1}^{h}$ with $\mathcal{W}_{\bullet}^{(j)}\in\mathbb{R}^{d_s\times d_h\times p}$; output weight $\mathcal{W}_O\in\mathbb{R}^{d_s\times d_s\times p}$; invertible transform $Z\in\mathbb{R}^{p\times p}$. \ENSURE Output $\mathcal{Y}\in\mathbb{R}^{T\times d_s\times p}$. \STATE $\mathcal{X}_{\text{pos}}\gets \mathcal{X}+\mathcal{P}$ \STATE $\widehat{\mathcal{X}}\gets \mathcal{X}_{\text{pos}}\times_3 Z$ \FOR{$j=1$ to $h$} \STATE $\widehat{\mathcal{W}}_Q^{(j)}\gets \mathcal{W}_Q^{(j)}\times_3 Z$ \STATE $\widehat{\mathcal{W}}_K^{(j)}\gets \mathcal{W}_K^{(j)}\times_3 Z$ \STATE $\widehat{\mathcal{W}}_V^{(j)}\gets \mathcal{W}_V^{(j)}\times_3 Z$ \ENDFOR \STATE $\widehat{\mathcal{W}}_O\gets \mathcal{W}_O\times_3 Z$ \FOR{$i=1$ to $p$} \STATE \textit{(loop over transform-domain slices; parallelizable)} \STATE $\widehat{X}^{(i)}\gets \widehat{\mathcal{X}}(:,:,i)\in\mathbb{R}^{T\times d_s}$ \FOR{$j=1$ to $h$} \STATE $\widehat{W}^{(j)}_{Q,i}\gets \widehat{\mathcal{W}}_Q^{(j)}(:,:,i)\in\mathbb{R}^{d_s\times d_h}$ \STATE $\widehat{W}^{(j)}_{K,i}\gets \widehat{\mathcal{W}}_K^{(j)}(:,:,i)\in\mathbb{R}^{d_s\times d_h}$ \STATE $\widehat{W}^{(j)}_{V,i}\gets \widehat{\mathcal{W}}_V^{(j)}(:,:,i)\in\mathbb{R}^{d_s\times d_h}$ \STATE $\widehat{Q}^{(j)}_i\gets \widehat{X}^{(i)}\,\widehat{W}^{(j)}_{Q,i}\in\mathbb{R}^{T\times d_h}$ \STATE $\widehat{K}^{(j)}_i\gets \widehat{X}^{(i)}\,\widehat{W}^{(j)}_{K,i}\in\mathbb{R}^{T\times d_h}$ \STATE $\widehat{V}^{(j)}_i\gets \widehat{X}^{(i)}\,\widehat{W}^{(j)}_{V,i}\in\mathbb{R}^{T\times d_h}$ \STATE $\widehat{S}^{(j)}_i\gets \frac{1}{\sqrt{d_h}}\,\widehat{Q}^{(j)}_i(\widehat{K}^{(j)}_i)^\top\in\mathbb{R}^{T\times T}$ \STATE $\widehat{A}^{(j)}_i\gets \mathrm{Softmax}(\widehat{S}^{(j)}_i)\in\mathbb{R}^{T\times T}$ \STATE $\widehat{\mathrm{head}}^{(j)}_i\gets \widehat{A}^{(j)}_i\,\widehat{V}^{(j)}_i\in\mathbb{R}^{T\times d_h}$ \ENDFOR \STATE $\widehat{H}_i\gets \mathrm{Concat}(\widehat{\mathrm{head}}^{(1)}_i,\dots,\widehat{\mathrm{head}}^{(h)}_i)\in\mathbb{R}^{T\times d_s}$ \STATE $\widehat{W}_{O,i}\gets \widehat{\mathcal{W}}_O(:,:,i)\in\mathbb{R}^{d_s\times d_s}$ \STATE $\widehat{Y}_i\gets \widehat{H}_i\,\widehat{W}_{O,i}\in\mathbb{R}^{T\times d_s}$ \ENDFOR \STATE Form $\widehat{\mathcal{Y}}\in\mathbb{R}^{T\times d_s\times p}$ by stacking $\{\widehat{Y}_i\}_{i=1}^p$ as frontal slices \STATE $\mathcal{Y}\gets \widehat{\mathcal{Y}}\times_3 Z^{-1}$ \RETURN $\mathcal{Y}$ \end{algorithmic} \end{algorithm}
\clearpage
\subsection{Pipeline view of slice-wise equivalence}
\label{app:pipeline}
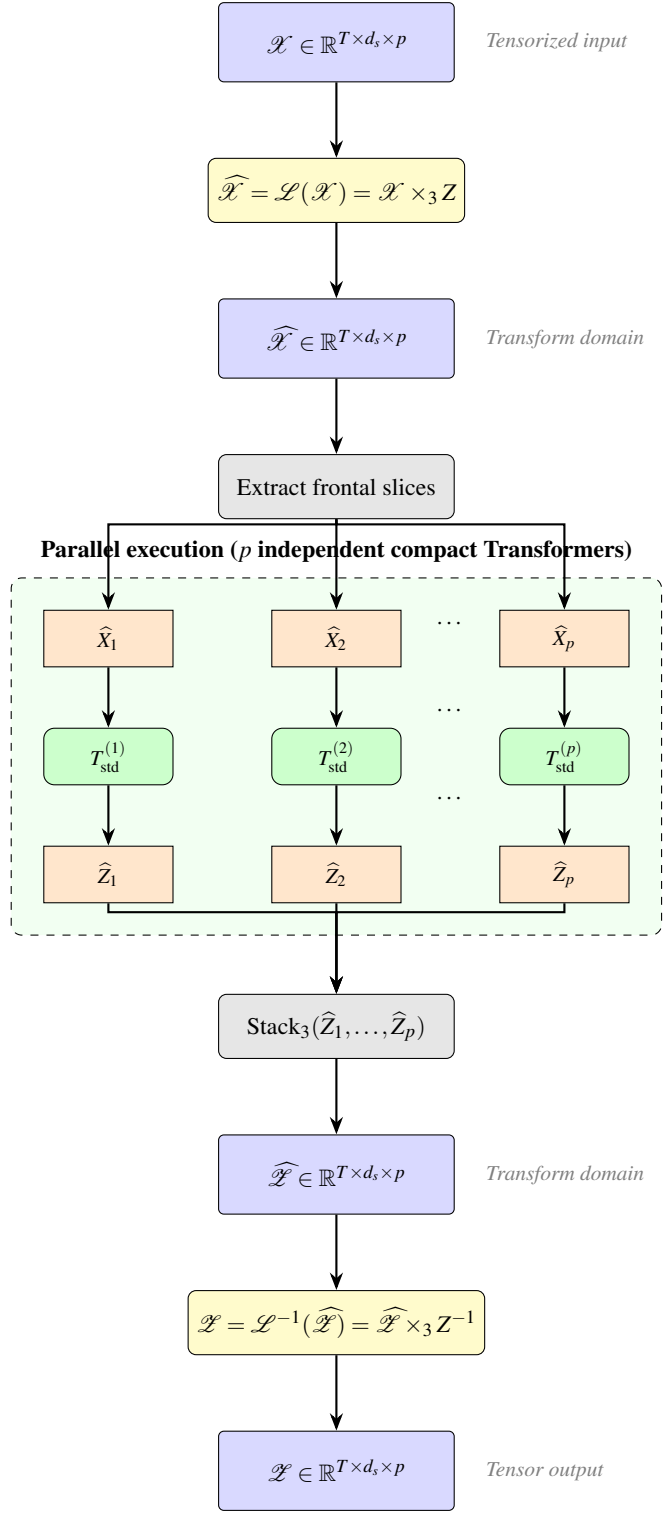
\begin{figure}[h!]
\small
\centering
\begin{tikzpicture}[
    >=Stealth,
    node distance=1.2cm,
    tensor/.style={draw, fill=blue!15, minimum width=3.1cm, minimum height=1.05cm, align=center, font=\small, rounded corners=2pt},
    transform/.style={draw, fill=yellow!25, minimum width=3.1cm, minimum height=0.85cm, align=center, font=\small, rounded corners=3pt},
    slice/.style={draw, fill=orange!20, minimum width=1.7cm, minimum height=0.75cm, align=center, font=\footnotesize},
    tblock/.style={draw, fill=green!20, rounded corners, minimum width=1.7cm, minimum height=0.75cm, align=center, font=\footnotesize},
    stack/.style={draw, fill=gray!20, minimum width=3.1cm, minimum height=0.85cm, align=center, font=\small, rounded corners=3pt},
    arrow/.style={->, thick},
    note/.style={font=\footnotesize\itshape, text=gray}
]
\node[tensor] (X) {$\mathcal{X}\in\mathbb{R}^{T\times d_s\times p}$};
\node[note, right=0.3cm of X] {Tensorized input};

\node[transform, below=1cm of X] (L) {$\widehat{\mathcal{X}}=\mathcal{L}(\mathcal{X})=\mathcal{X}\times_3 Z$};

\node[tensor, below=1cm of L] (Xhat) {$\widehat{\mathcal{X}}\in\mathbb{R}^{T\times d_s\times p}$};
\node[note, right=0.3cm of Xhat] {Transform domain};

\node[stack, below=1cm of Xhat] (extract) {Extract frontal slices};

\node[slice, below=1.2cm of extract, xshift=-3.0cm] (X1) {$\widehat{X}_1$};
\node[slice, below=1.2cm of extract, xshift=0cm] (X2) {$\widehat{X}_2$};
\node[slice, below=1.2cm of extract, xshift=3.0cm] (Xp) {$\widehat{X}_p$};
\node[below=1.2cm of extract, xshift=1.5cm] (dots1) {$\cdots$};

\node[tblock, below=0.8cm of X1] (T1) {$T_{\mathrm{std}}^{(1)}$};
\node[tblock, below=0.8cm of X2] (T2) {$T_{\mathrm{std}}^{(2)}$};
\node[tblock, below=0.8cm of Xp] (Tp) {$T_{\mathrm{std}}^{(p)}$};
\node[below=0.8cm of dots1] (dots2) {$\cdots$};

\node[slice, below=0.8cm of T1] (Z1) {$\widehat{Z}_1$};
\node[slice, below=0.8cm of T2] (Z2) {$\widehat{Z}_2$};
\node[slice, below=0.8cm of Tp] (Zp) {$\widehat{Z}_p$};
\node[below=0.8cm of dots2] (dots3) {$\cdots$};

\node[stack, below=1.2cm of Z2] (Stack) {$\text{Stack}_3(\widehat{Z}_1,\dots,\widehat{Z}_p)$};

\node[tensor, below=1cm of Stack] (Zhat) {$\widehat{\mathcal{Z}}\in\mathbb{R}^{T\times d_s\times p}$};
\node[note, right=0.3cm of Zhat] {Transform domain};

\node[transform, below=1cm of Zhat] (Linverse) {$\mathcal{Z}=\mathcal{L}^{-1}(\widehat{\mathcal{Z}})=\widehat{\mathcal{Z}}\times_3 Z^{-1}$};

\node[tensor, below=1cm of Linverse] (Z) {$\mathcal{Z}\in\mathbb{R}^{T\times d_s\times p}$};
\node[note, right=0.3cm of Z] {Tensor output};

\draw[arrow] (X) -- (L);
\draw[arrow] (L) -- (Xhat);
\draw[arrow] (Xhat) -- (extract);

\draw[arrow] (extract) -- ++(0,-0.5) -| (X1);
\draw[arrow] (extract) -- (X2);
\draw[arrow] (extract) -- ++(0,-0.5) -| (Xp);

\draw[arrow] (X1) -- (T1);
\draw[arrow] (X2) -- (T2);
\draw[arrow] (Xp) -- (Tp);

\draw[arrow] (T1) -- (Z1);
\draw[arrow] (T2) -- (Z2);
\draw[arrow] (Tp) -- (Zp);

\draw[arrow] (Z1) -- ++(0,-0.5) -| (Stack);
\draw[arrow] (Z2) -- (Stack);
\draw[arrow] (Zp) -- ++(0,-0.5) -| (Stack);

\draw[arrow] (Stack) -- (Zhat);
\draw[arrow] (Zhat) -- (Linverse);
\draw[arrow] (Linverse) -- (Z);

\begin{scope}[on background layer]
\node[draw, dashed, rounded corners, fit=(X1)(X2)(Xp)(T1)(T2)(Tp)(Z1)(Z2)(Zp), inner sep=12pt, fill=green!5] (parallel) {};
\end{scope}
\node[above=0.1cm of parallel.north, font=\small\bfseries] {Parallel execution ($p$ independent compact Transformers)};
\end{tikzpicture}
\caption{Slice-wise equivalence of the tensor Transformer under the $\mathcal{L}$-product framework.
The tensor is transformed along mode-3, processed independently slice-by-slice by $p$ compact Transformers (dimension $d_s$), stacked back, and mapped to the original domain via $\mathcal{L}^{-1}$.}
\label{fig:pipeline}
\end{figure}

Figure~\ref{fig:pipeline} provides an implementation-oriented view of the slice-wise equivalence result by
making the dataflow explicit. One tensor encoder block can be understood as a three-step pipeline:
(i) apply the $\mathcal{L}$-transform along the tube (mode-3) dimension to obtain $p$ transform-domain
frontal slices $\{\widehat{X}_i\}_{i=1}^p$, (ii) process these slices independently using $p$ compact
standard Transformer blocks $T_{\mathrm{std}}^{(i)}$ of width $d_s$ (which can be executed in parallel or
via batched kernels), and (iii) stack the slice outputs and apply $\mathcal{L}^{-1}$ to return to the
original tensor domain. This diagram therefore clarifies both the source of the $\approx 1/p$ encoder
parameter scaling (each sub-encoder operates on width $d_s=d/p$) and the practical parallelism available
in the transform domain. Importantly, the model is not a static partition of the embedding dimension:
after each block, the inverse transform $\mathcal{L}^{-1}$ linearly re-mixes the $p$ spectral channels,
so the slices fed into the next layer are coupled combinations of the previous layer outputs, enabling
cross-slice information flow across depth even though the within-block computation is slice-wise in the
$\mathcal{L}$-domain.

\subsection{Remark on FFT/t-product and complex-valued slices}
\label{app:complex}

This remark clarifies what changes (and what does not) if one instantiates the $\mathcal{L}$-product with a Fourier
transform, i.e., the classical t-product setting.

\paragraph{Complex transform-domain representations.}
When $Z$ is chosen as the DFT matrix, the $\mathcal{L}$-transform $\widehat{\mathcal{X}}=\mathcal{X}\times_3 Z$
is generally \emph{complex-valued} even when $\mathcal{X}$ is real. Consequently, the slice-wise ``compact Transformer''
interpretation remains valid, but all linear algebra must be understood over $\mathbb{C}$ and inner products must use
the Hermitian form.

\paragraph{Attention in $\mathbb{C}$ is well-defined (with Hermitian transpose).}
Within each slice $i$, attention scores are computed using conjugate transpose:
\[
\widehat{S}_i^{(j)}=\frac{1}{\sqrt{d_h}}\widehat{Q}_i^{(j)}(\widehat{K}_i^{(j)})^{H},
\]
and the remaining operations (softmax, weighting by $\widehat{V}_i^{(j)}$, output projection) proceed slice-wise as in
the real case. The resulting $\widehat{\mathcal{Y}}$ is complex in general, and the inverse transform
$\mathcal{L}^{-1}(\widehat{\mathcal{Y}})$ recovers a real tensor only when the appropriate conjugate-symmetry
constraints hold (as in standard FFT-based real-signal processing).

Although FFT/t-product choices are theoretically natural and computationally efficient, they introduce practical
complications in typical deep-learning stacks: complex-valued parameters/activations, potential ambiguity about the
softmax on complex scores (often handled by taking real parts or magnitudes), and additional care to preserve
real-valued outputs. For these reasons, we adopt \emph{real orthogonal transforms} (notably DCT-type transforms
\cite{Ahmed1974DCT}) so that all activations and parameters remain in $\mathbb{R}$, the attention computation matches
standard real-valued implementations, and the induced spectral slices admit a direct interpretation as real channels.
This choice keeps the method fully compatible with standard training pipelines while retaining the key algebraic
property needed for slice-wise decoupling in the transform domain.
\section{Training curves}
\label{app:learning_curves}
This appendix complements Section~\ref{sec:setup} and Section~\ref{sec:main_results} by showing the
\emph{optimization dynamics} of the baseline and tensorized encoders.
While the main text reports end-point performance (accuracy/F1) aggregated over seeds, the curves help verify that
(i) the reported improvements are not due to unstable training, and (ii) spectral weighting mainly affects convergence
behaviour rather than creating erratic runs.
All curves are shown for a representative seed (seed~42); the remaining seeds exhibit the same qualitative trends.

Each figure reports the evolution across epochs of the main evaluation metrics (accuracy and macro-F1) together with the loss.
The IMDB experiment uses 20 epochs, while AG~News ($d=256$) uses 5 epochs, matching the schedules described in
Table~\ref{tab:train_by_task}.

\begin{figure}[h!]
  \centering
  \includegraphics[width=\textwidth]{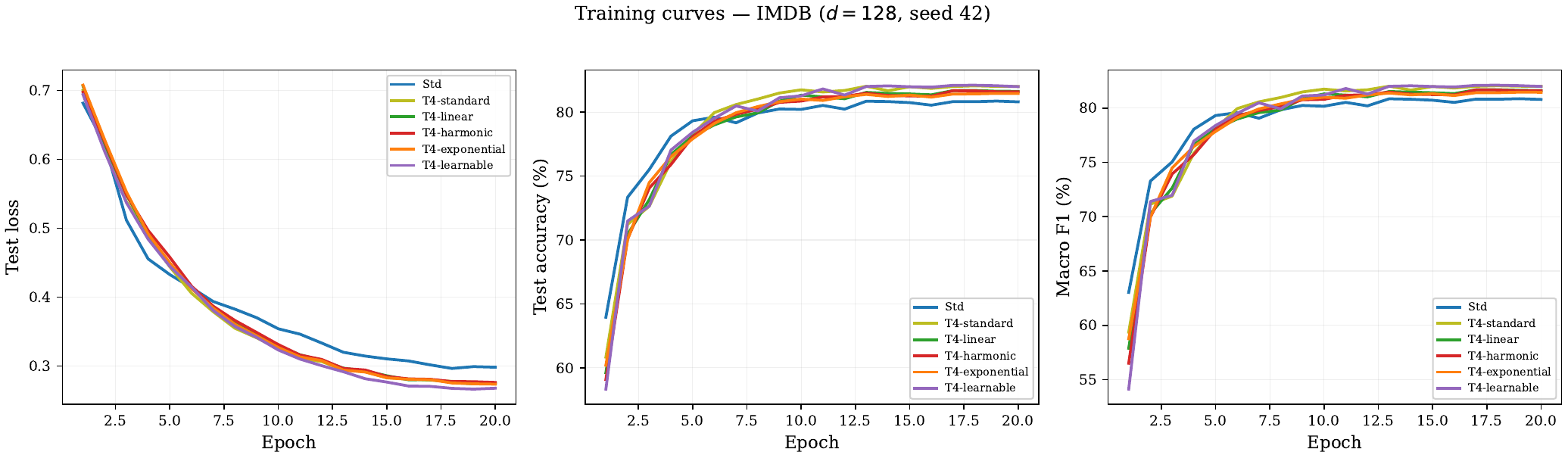}
  \caption{Training curves on IMDB ($d=128$, seed 42).}
  \label{fig:lc_imdb}
\end{figure}
Across PE strategies, tensor models reach a stable high-accuracy regime quickly and maintain it with limited oscillations,
consistent with the low seed-to-seed variance reported in Table~\ref{tab:main_imdb}.
Moreover, the relative ordering between PE strategies observed in the final metrics is already visible early in training,
suggesting that the gains are not due to late-epoch noise.
\begin{figure}[h]
  \centering
  \includegraphics[width=\textwidth]{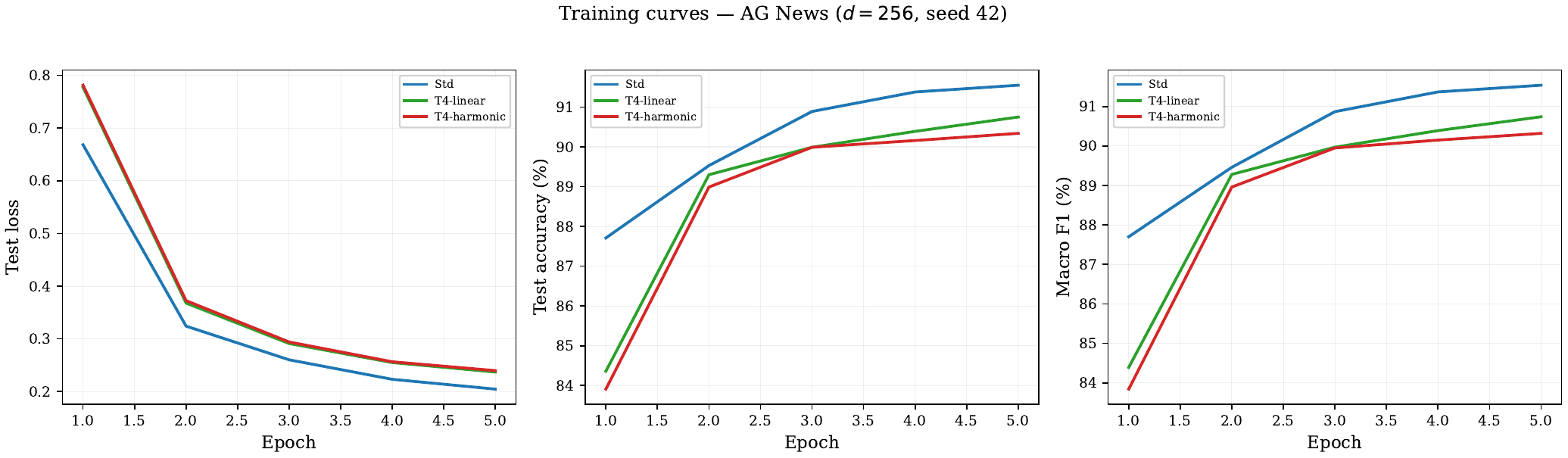}
  \caption{Training curves on AG~News ($d=256$, seed 42).}
  \label{fig:lc_ag256}
\end{figure}
\noindent
All models peak within the short 5-epoch budget, leaving less room for dramatic convergence differences than IMDB.
The tensor variants remain stable, and the modest accuracy gap reported in Table~\ref{tab:main_agnews} is consistent with the
curves: the tensor model trades a small amount of accuracy for a large reduction in encoder parameters at this width.

\section{Full PE sweep}
\label{app:pe_sweep}
Table~\ref{tab:pe_full} reports the full comparison of the five spectral weighting schemes at fixed $p=4$.
The purpose is twofold: (i) to support the main-text claim that \emph{spectral weighting matters but is not brittle},
and (ii) to document the relative ranking of strategies beyond the single figure shown in the main paper.

Each entry is the mean test accuracy $\pm$ standard deviation over 3 seeds, and ``Std'' denotes the full-width baseline.
Boldface indicates the best tensor strategy within each dataset among the runs reported here.
Missing entries should be interpreted as ``not run / not available'' (and should be completed if possible for full coverage). 

\begin{table}[h!]
  \centering
  \caption{Full PE sweep at $p=4$ (mean $\pm$ std, 3 seeds). Best per dataset in \textbf{bold}.}
  \label{tab:pe_full}
  \small
  \begin{tabular}{lcc}
    \toprule
    \textbf{PE strategy} & \textbf{IMDB $d=128$} & \textbf{AG News $d=256$} \\
    \midrule
    T4-standard ($\alpha_k=1$) & $\mathbf{82.02 \pm 0.03}$ & --- \\
    T4-linear ($\alpha_k=k/p$) & $81.65 \pm 0.06$ & $\mathbf{90.76 \pm 0.13}$ \\
    T4-exponential ($\alpha_k=2^{(k-1)/(p-1)}$) & $81.60 \pm 0.14$ & --- \\
    T4-harmonic ($\alpha_k=k$) & $81.53 \pm 0.13$ & $90.41 \pm 0.08$ \\
    T4-learnable & $81.98 \pm 0.20$ & --- \\
    \midrule
    Std (baseline) & $80.77 \pm 0.24$ & $91.40 \pm 0.31$ \\
    \bottomrule
  \end{tabular}
\end{table}
On IMDB, all PE strategies outperform Std, and the spread across strategies remains under one percentage point,
indicating that the tensorization is the dominant driver while weighting provides a secondary tuning knob.
On AG~News at $d=256$, linear weighting performs best among the reported tensor runs.
Overall, the results support the main-text conclusion that T4-linear is a robust default, while the best strategy can remain dataset-dependent.

\section{Runtime details}
\label{app:runtime}
Table~\ref{tab:runtime_full} provides wall-clock and memory measurements that complement the FLOPs analysis in
Section~\ref{sec:efficiency}.
The goal is to clarify (i) what is reduced by the tensor factorization (encoder parameters and encoder FLOPs),
and (ii) what is currently limited by implementation details (sequential slice execution).

``Train'' is the mean wall-clock time per epoch, ``Mem'' is the peak GPU memory allocation during the run,
and ``Infer'' is the wall-clock time of a forward pass over the full test set.
All measurements are taken on the same GPU (Tesla T4) with mixed precision.
Because AG~News at $d=768$ uses a different batch size than smaller configurations, absolute inference times
should be compared primarily within the same configuration.
\begin{table}[h!]
  \centering
  \caption{Runtime details. Train is mean wall-clock per epoch; Mem is peak GPU allocation; Infer is forward pass on the full test set. All runs on NVIDIA Tesla T4 with mixed precision.}
  \label{tab:runtime_full}
  \small
  \setlength{\tabcolsep}{4pt}
  \renewcommand{\arraystretch}{1.05}
  \begin{tabular}{p{2.1cm} l c c c c}
    \toprule
    \textbf{Config} & \textbf{Model} & \textbf{Enc.} & \textbf{Train (s/ep)} & \textbf{Mem (GB)} & \textbf{Infer (s)} \\
    \midrule
    \multirow{6}{=}{IMDB ($d=128$)}
      & Std           & 991\,K & 26  & 0.50 & 11.8 \\
      & T4-standard   & 254\,K & 34  & 0.51 & 12.7 \\
      & T4-linear     & 254\,K & 34  & 0.51 & 12.6 \\
      & T4-harmonic   & 254\,K & 34  & 0.51 & 12.8 \\
      & T4-learnable  & 270\,K & 34  & 0.51 & 12.8 \\
      & T2-linear     & 500\,K & 29  & 0.49 & 11.9 \\
    \midrule
    \multirow{3}{=}{AG~News ($d=256$)}
      & Std           & 3.95\,M & 87  & 0.99 & 3.0 \\
      & T4-linear     & 1.00\,M & 116 & 0.96 & 3.3 \\
      & T4-harmonic   & 1.00\,M & 116 & 0.96 & 3.2 \\
    \midrule
    \multirow{2}{=}{AG~News ($d=768$)}
      & Std           & 28.4\,M & 260 & 1.95 & 5.0 \\
      & T4-linear     & 7.1\,M  & 241& 1.66 & 5.2\\
    \bottomrule
  \end{tabular}
\end{table}

The tensor encoder achieves the expected parameter/FLOPs reductions, but the current implementation executes the
$p$ sub-encoders sequentially, which increases epoch time at smaller widths (IMDB and AG $d=256$).
In contrast, peak memory is already improved at large width ($d=768$), where the tensor model reduces allocation by 15\%.
This points to a clear engineering lever: batching slices (treating the slice index as a batch dimension) should allow the
FLOPs reduction to translate into wall-clock improvements, especially at larger widths where compute dominates overhead.

\paragraph{Wall-clock overhead.}
The current tensor implementation processes the $p$ sub-encoders sequentially, which prevents full exploitation of the $1/p$ FLOPs reduction in wall-clock time.
On IMDB, this corresponds to 34 seconds per epoch versus 26 seconds for Std.
On AG~News at $d=256$, this corresponds to 116 seconds per epoch versus 87 seconds for Std.

\paragraph{Peak memory.}
Peak memory is comparable at moderate widths and improves at large width.
At $d=768$, the tensor model reduces peak allocation by 15\% (1.66 vs 1.95 GB), which is practically meaningful under tight GPU memory budgets.

\paragraph{Batched-slice potential.}
A batched-slice implementation (treating the slice dimension as a batch dimension) would eliminate the sequential overhead.
Given the $4\times$ FLOPs reduction in the encoder, batched execution is expected to match or improve upon Std wall-clock time at moderate to large widths.

\clearpage
\bibliographystyle{plain}  
\bibliography{bibliography} 
\end{document}